\documentclass[preprint,12pt]{elsarticle}

\usepackage{amsthm,amsmath,amssymb}
\usepackage[colorlinks=true,linkcolor=blue,citecolor=blue,urlcolor=blue,]{hyperref}
\usepackage{cleveref}
\usepackage{booktabs}
\usepackage{multirow}
\usepackage{makecell}
\usepackage{dblfloatfix}

\journal{arvix}

\begin{document}

\begin{frontmatter}

\title{AGMR-Net: Attention Guided Multiscale Recovery framework for stroke segmentation}
\author{Xiuquan Du\corref{cor1}\fnref{label1,label2}}
\ead{dxqllp@163.com}
\cortext[cor1]{Corresponding author}
\author[label2]{Kunpeng Ma}\ead{2297595655@qq.com}
\author[label2]{Yuhui Song}\ead{leesingle1993@gmail.com}


\address[label1]{Key Laboratory of Intelligent Computing and Signal Processing of Ministry of Education, Anhui University, Hefei 230601, Anhui, China }
\address[label2]{School of Computer Science and Technology, Anhui University, Hefei 230601, Anhui, China}



\begin{abstract}
Automatic and accurate lesion segmentation is critical for clinically estimating the lesion statuses of stroke diseases and developing appropriate diagnostic systems. Although existing methods have achieved remarkable results, further adoption of the models is hindered by: (1) inter-class indistinction, the normal brain tissue resembles the lesion in appearance. (2) intra-class inconsistency, large variability exists between different areas of the lesion.  To solve these challenges in stroke segmentation, we propose a novel method, namely Attention Guided Multiscale Recovery framework (AGMR-Net) in this paper. Firstly, a coarse-grained patch attention module in the encoding is adopted to get a patch-based coarse-grained attention map in a multi-stage explicitly supervised way, enabling target spatial context saliency representation with a patch-based weighting technique that eliminates the effect of intra-class inconsistency. Secondly, to obtain a more detailed boundary partitioning to solve the challenge of the inter-class indistinction, a newly designed cross-dimensional feature fusion module is used to capture global contextual information to further guide the selective aggregation of 2D and 3D features, which can compensate for the lack of boundary learning capability of 2D convolution. Lastly, in the decoding stage, an innovative designed multi-scale deconvolution upsampling  instead of linear interpolation enhances the recovery of target space and boundary information. The AGMR-Net is evaluated on the open dataset Anatomical Tracings of Lesions-After-Stroke (ATLAS), achieving the highest dice similarity coefficient (DSC) score of 0.594, Hausdorff distance of 27.005 mm, and average symmetry surface distance of 7.137 mm, which demonstrate that our proposed method outperforms other state-of-the-art methods and has great potential in the diagnosis of stroke.

\end{abstract}




\end{frontmatter}


\section{Introduction}

\begin{figure*}[!t]
\centering
\includegraphics[width=0.9\textwidth]{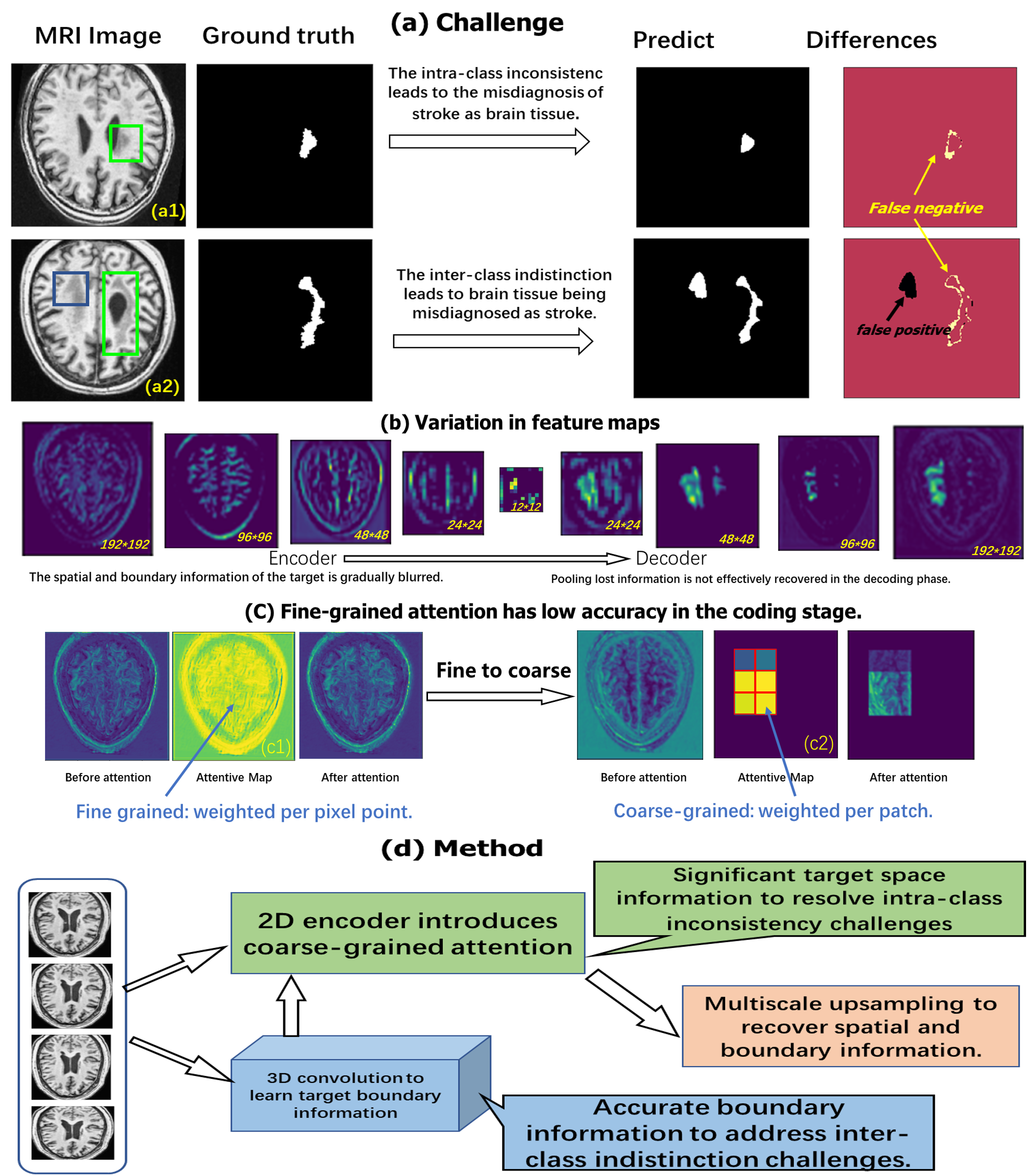}
\caption{(a): Examples of tissue and lesion similarity.  (b): The ideal fine-grained attention. (c): The ideal coarse-grained attention. (d): A map of the features after four pooling.}
\label{fig:1}
\end{figure*}

Stroke is ischemia due to rupture or congestion of blood vessels in the brain \cite{1}. In clinical practice, physicians need to accurately map out stroke lesions, which are used to calculate the size, shape, and volume of the lesion. This task is not only time-consuming but also relies on the subjective judgment of physicians, which is not suitable for larger data. There is a clear need for an automatic method to help physicians segment lesion regions. With recent developments in full convolutional networks (FCN), particularly convolutional neural networks(CNN), many great works \cite{4,5,6,7} have been done in stroke segmentation with good results. However, the features learned by these methods still face some challenges, as shown in Fig. \ref{fig:1}.(a): (1)intra-class inconsistent. Since the network lacks guided learning of the target space information, only some of the lesions in the green boxes in
 Fig. \ref{fig:1}.(a1) are identified, and there are still some regions that are misclassified as brain tissue (false negatives) because of significant differences between them and the lesions. (2) inter-class indistinction, as shown in Fig. \ref{fig:1}.(a2), where the brain tissue in the blue box is very similar to the lesion. The network is limited by the precise target boundary delineation of the interclass feature, which misclassifies brain tissue as stroke (false positive).

Significant target spatial information can guide the network to mine features that are specific to the target correlation, and it will overcome the drawback of inter-class inconsistency.   As shown in Fig. \ref{fig:1}.(b), the spatial information (location information) of the target is less significant in the network encoding stage than in the decoding stage, and some works have attempted to represent the target spatial information significantly in the encoding stage by employing fine-grained spatial attention \cite{8,9,10,11} methods. However, as shown in 
Fig. \ref{fig:1}.(c1), the resulting fine-grained attention map enhances all regions indiscriminately. This is because (1) there is insufficient learning information to produce an accurate fine-grained attention map while the encoding phase learns some detailed features of the target, such as edges or corners. (2) In the encoding phase, with the pooling operation, as indicated in Fig. \ref{fig:1}.(b), as the feature map grows more abstract and the target features become too unpredictable to incorporate prior information containing spatial information, the accuracy of salient regions cannot be guaranteed by explicit training and supervision \cite{26}. The lack of attention accuracy limits its usefulness in segmentation \cite{41}. For this reason, PDNet \cite{43} introduced a rough method for extracting lesion regions by drawing a circle covering the target with the click information as the center and monitoring the diameter of the circle to control the accuracy of attention. However, the click information as an input needs to be marked manually by the physician. Therefore, it is necessary to design a coarse-grained attention module that can be supervised without physician labeling to guide the network to focus on the target region.

In addition, the accurate representation of boundary features can amplify the differences in features between classes \cite{60} and helps to solve the challenge of inter-class indistinction. Previous work \cite{24,25} used 3D convolution to compensate for the lack of 2D convolution, which allows for more accurate boundary segmentation. Unfortunately, the naive behavior of these methods to aggregate features by direct summation or concatenation, it perhaps produces redundant information and noise that corrupts the original features \cite{7}. Further, D-UNnet \cite{31} introduced two parallel designed Squeeze-and-Excitation \cite{14} mechanisms for 2D and 3D feature fusion, filtering and screening different dimensional features for fusion to achieve better fusion results. However, in the fusion process,  it is the global features of two dimensions are brought together, but the relationship between the modeled global channels is ignored to model only the local relationship of the feature channels in a single dimension, which is still an obstacle to the global feature fusion between different dimensions. We believe that the global dependencies of the feature channels between different dimensions need to be considered, and it will be more beneficial for feature fusion.

Nevertheless,as indicated in Fig. \ref{fig:1}.(b), the effective target boundary and spatial information that is valuable for segmentation in the encoding phase is potentially lost in subsequent pooling operations or convolutional processes to reduce the resolution. It will impede medical picture segmentation with demanding prediction tasks, necessitating the recovery of more spatial and boundary information during the decoding step to improve segmentation performance. A more appropriate feature fusion in the skip-connected encoding and decoding stages \cite{4,6,17} can compensate for the lost data. However, insufficient attention has been devoted to feature enhancement during upsampling, and basic parameterless linear interpolation of upsampling is deleterious \cite{15}.

To face the above problems, in this study we propose the Attention Guided Multiscale Recovery framework (AGMR-Net), As shown in Fig. \ref{fig:1}.(d), our basic idea is to first design coarse-grained attentional salient target spatial information in the encoding stage to solve the intra-class inconsistency, then introduce precise boundary information learned by 3D convolution to solve the challenge of inter-class indistinction, and finally utilize multi-scale information to recover the loss of spatial and boundary information caused by pooling. Specifically, AGMR-Net uses 2D and 3D convolution for stroke feature extraction in the encoding phase. To facilitate the efficient extraction of focal target features in the encoding stage, a novel coarse-grained patch attention module (CPA) is designed to generate a patch-based coarse-grained attention map to highlight target spatial information in a coarse-grained weighted manner. As shown in Fig. \ref{fig:1}.(c2), each red patch has a target on the way, and we enhance each target patch with coarse-grained attention at the patch level, which can guide the network to focus more on extracting target features and solve the problem of inter-class indistinction. In addition, a prior significant map is introduced into the CPA as the prior knowledge makes attention more accurate and the CPA easy to be trained. After that, a cross-dimensional feature fusion module (CFF) module modeling global channel dependencies is designed to adaptively fuse 2D and 3D features to refine the boundary representation of 2D features and address the challenge of intra-class inconsistency. However, it is limited by the information loss caused by pooling operations. Finally, an innovative multi-scale deconvolution upsampling module (MDU) uses multi-scale feature information to recover the target space information and boundary information lost by pooling operations.

The contributions of this work are summarized as follows:
\begin{itemize}
\item The novel CPA extracts coarse-grained patch-based attention to enhance target feature extraction. In addition, the prior significant map, which is the a priori knowledge of CPA, facilitates the training of CPA and ensures the accuracy of the generated attention map. 
\item The innovative CFF proposes a global information guide incorporating multidimensional features to refine the boundary feature representation in the encoding phase.
\item The newly designed MDU replaces the parameter-free linear interpolation to recover the target spatial information and boundary information lost by pooling in the encoding stage by capturing the multiscale information.
\end{itemize}

\section{Related Works}
Recently, many encoder-decoder-based approaches \cite{6,26,5} have achieved excellent performance in segmentation tasks. Most of them still suffer from intra-class inconsistency and inter-class indistinction challenges. In this section, we briefly describe spatial attention methods that enable significant target spatial information representation in encoders and 2D-3D features fusion methods that refine target boundary features, as well as methods that recover spatial information and boundary detail information in decoders. And beyond that, we also summarize the forms of related methods used and their advantages in Table \ref{tab:1}.

\begin{table}[h]
\caption{Summary of related work.}
\resizebox{\linewidth}{!}{
\begin{tabular}{ccccp{5cm}p{8cm}}
\hline
Methods                      & Author                                                                       & Dataset & Attentional style         & Advantages                                                & Disadvantages                                                                                       \\ \hline
Attention module             & \begin{tabular}[c]{@{}l@{}}Liangliang Liu’s Work \cite{6}\end{tabular} & 3D      & fine-grained              & Significant spatial information                           & No explicit training                                                                                \\
patch-based attention        & Nils Gessert’s   Work \cite{44}                                                 & 2D      & coarse-grained            & Flexible and easy   to embed                              & No explicit   training                                                                              \\
Context Prior                & Changqian Yu’s Work \cite{26}                                                  & 2D      & fine-grained              & Explicit training                                         & Only a single stage can be supervised.                                                               \\
Click-driven Attention       & Y Tang’s Work \cite{43}                                                       & 3D      & coarse-grained            & Explicit training                                         & A priori   information needs to be marked manually.                                                  \\ \hline
Methods                      & Author                                                                       & Dataset & Aggregation   method      & Advantages                                                & Disadvantages                                                                                       \\ \hline
Progressive Fusion Network   & Chaowei Fang’s Work \cite{24}                                                     & 3D      & Concatenate               & Ability to learn 3D complex set information.              & Directly fusing the features of the two dimensions introduces noise.                                \\
Dimension Transform Block    & D-Unet’s Work \cite{31}                                                          & 3D      & Adaptive Screening Fusion & Feature screening is performed before feature fusion.     & Global information between different tasks is not   considered.                                     \\ \hline
Methods                      & Author                                                                        & Dataset & Used stage                & Advantages                                                & Disadvantages                                                                                       \\ \hline
\multirow{2}{*}{ASPP module} & Guotai Wang’s Work \cite{18}                                                      & 3D      & Encoding stage            & \multirow{2}{*}{\makecell[c]{Capture spatial and \\ boundary information.}} & \multirow{2}{*}{\makecell[c]{Ignores recovery space and \\ boundary information during upsampling.}} \\
                             & Hao Yan’s Work \cite{4}                                                         & 3D      & Encoding stage            &                                                           &                                                                                                     \\ \hline
\end{tabular}}
\label{tab:1}
\end{table}

\subsection{ Spatial attention method}
In order to significantly the target spatial information, the network is guided to extract the target features. CBAM \cite{30} developed and validated attention module, pooling techniques along the channel axis are used to construct the spatial attention maps. The spatial domain attention \cite{8,9,10,11} mechanism has been widely applied in medical image segmentation, Table \ref{tab:1} lists coarse-grained, fine-grained, supervised, and unsupervised spatial attention methods, demonstrating the effectiveness of spatial attention for medical image segmentation. For example, DRANet \cite{6} adopted an unsupervised fine-grained attention approach that enables CNN to extract target-specific features, hence improving segmentation performance. However, the encoder part of DRANet tends to learn detailed features and lacks regularization, which is insufficient to enable the network to create more accurate pixel-based fine-grained attention. In contrast to fine-grained attention using local attention obtained by convolution, further, Nils Gessert's work \cite{44} proposed coarse-grained attention methods, local and global information aggregation strategies, to learn the global context between coarse-grained patches. Although the results are promising and demonstrate the feasibility of coarse-grained attention, the accuracy of attention is still limited by the lack of explicit supervision.

Recently, some networks were proposed with a supervision mechanism, such as CPNet \cite{26}, which included an a prior knowledge inter-class affinity map to explicitly supervise and train to guarantee the accuracy of the attention graph generation. Unfortunately, it supervises only one stage of attention graphs. To ensure the correctness of multi-stage attention, PDNet \cite{43} used click information as input to guide the generation of the attention map, and generated click images and distance transformed images \cite{52} to introduce multi-stage explicit supervision to guarantee the accuracy of attention. However, their a priori click information requires manual annotation by a physician. Based on these studies, we propose a coarse-grained attention-based approach that can be trained by multi-layer supervision and the a prior information does not require manual annotation.  Although the above strategy aids the network in extracting target-specific characteristics, 2D convolution in volumetric data lacks significant boundary learning capabilities.

\subsection{ 2D-3D features fusion methods}
In 3D data segmentation of medical images, the accurate boundary information learned by 3D convolution can enrich the details of 2D convolutional feature boundaries and obtain better segmentation performance. ComboNet \cite{46} directly merged the segmentation results of 2D CNN and 3D CNN, which can retain the precision of the target border, but the 3D CNN is too computationally intensive. In order to solve this problem, some networks fuse 2D and 3D features only in the encoding phase to facilitate the interaction of feature complementarity while reducing the computational effort, e.g., M2D3DNet \cite{25}, GGPFN \cite{24}. Unfortunately, as shown in Table \ref{tab:1}, these naive connection methods do not sufficiently take into account dependencies between the characteristics of different tasks \cite{49}. 

Learning the dependency of each feature corresponding to a channel can emphasize task-relevant features  \cite{48,57}. SENet \cite{14} proposed the Squeeze-and-Excitation module, which selects different weights according to different parts of the relevant input of the task \cite{30}. Thus, D-Unet \cite{31} attempted the use of a channel focus approach to improving the feature fusion task to bridge the gap, greatly improving the network performance. However, it merely extracts the relationship between channels inside a dimension without taking into account global information between dimensions, thus dependencies between features in various dimensions are neglected. It is necessary to design a method to learn the global information of different dimensional features to guide the fusion.

\subsection{Recovery space and target information method}

In code-decoder structures for medical image segmentation, boundary and spatial information are lost in the down-sampling of the encoder, and recovering this information in the decoding can improve the segmentation. the skip connection of the U-Net \cite{12} structure is to reintroduce the information from the encoded features to the decoded features. There are also many works that focus on the effective fusion of encoder and decoder information, for example, Attention Unet \cite{34} employed a gating mechanism, and CLCI-Net \cite{18} proposed a convolutional LSTM \cite{51} to bring some detailed information from the encoder to the  decoder to compensate for information loss due to successive pooling and stepwise convolutional operations. However, the upsampling process of the encoder ignores spatial and boundary information acquisition when using basic linear interpolation \cite{25,31,12} and deconvolution \cite{50} to recover the resolution.

By capturing multi-scale characteristics, CNN-based techniques can recover rich spatial and boundary information about target features \cite{19}. Astrous spatial pyramid pooling (ASPP) module is proposed in Deeplab V3 \cite{16} to learn multi-scale contextual information. PSPNet \cite{17} used a pyramid pooling module (PPM) to enhance the network's ability to exploit global contextual information. The applications of multiscale methods in medical image segmentation are also listed in Table \ref{tab:1}. COPLE-Net \cite{18} used ASPP modules in the bottleneck region to enrich multiscale features, and CLCI-Net \cite{4} bundled feature maps from different stages of the network based on ASPP, and all these networks utilized multiscale information in the encoding stage and also prove their effectiveness.  We claim that employing parametric multiscale deconvolution to capture spatial and detailed information during the upsampling procedure is more advantageous for improving the segmentation task for dense prediction.

\section{Methods}
\begin{figure}[h]
    \centering
    \includegraphics[width=1\textwidth]{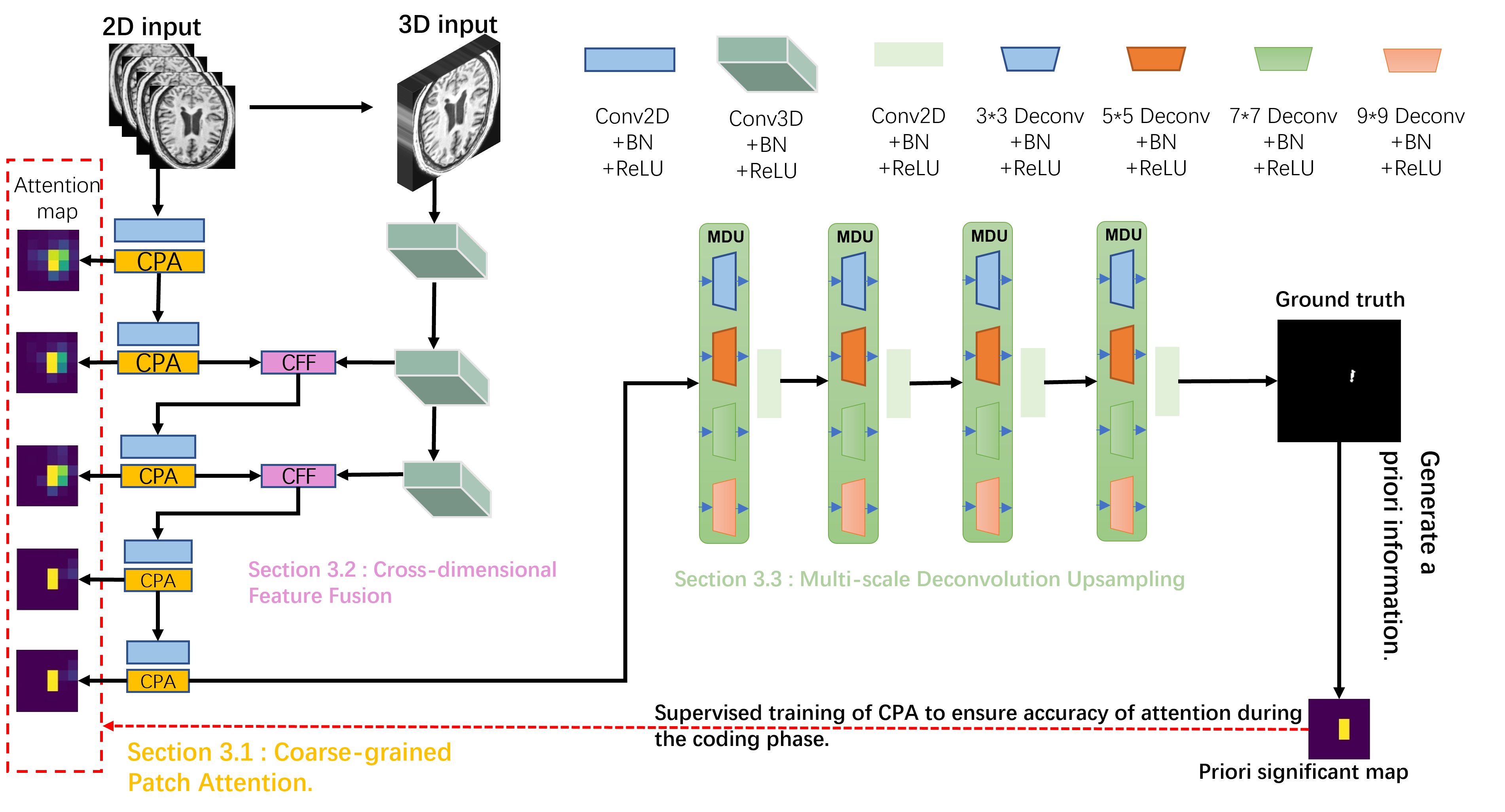} 
    \caption{Overview of the proposed AGMR-Net. Four consecutive slices are fed into the network, using a 3D encoder as a secondary network to supplement the 2D network with boundary information.}
    \label{fig:2}
\end{figure}

In this work, we propose a novel attention-guided multi-scale recovery framework, as shown in Fig. \ref{fig:2}. First, we introduce a coarse-grained patch attention module (CPA) that generates a coarse-grained attention graph that significantly represents the spatial information of the target solving the problem of inter-class similarity, and the attention graph generated based on the labels serves as prior knowledge to facilitate the training of the CPA. Secondly, to refine the boundary representation of these features, we propose a cross-dimensional feature fusion (CFF) module at the intersection of 2D and 3D coding layers and decide on the choice of complementary features that will facilitate segmentation. Finally, a multiscale deconvolution upsampling module (MDU) is designed to capture multiscale contextual information and recover target location information and boundary information lost after multiple pooling in the abstract feature map. Next, we describe these modules in detail.

\subsection{ Coarse-grained Patch Attention (CPA)}
In the 2D encoding layer of the network, we design coarse-grained patch attention with prior knowledge to enhance target cues in the feature map and reduce redundant noise information in the background, to generate high-quality features. Here we present the details of the coarse-grained patch attention (CPA) module.

\begin{figure*}[h]
    \centering
    \includegraphics[width=1\textwidth]{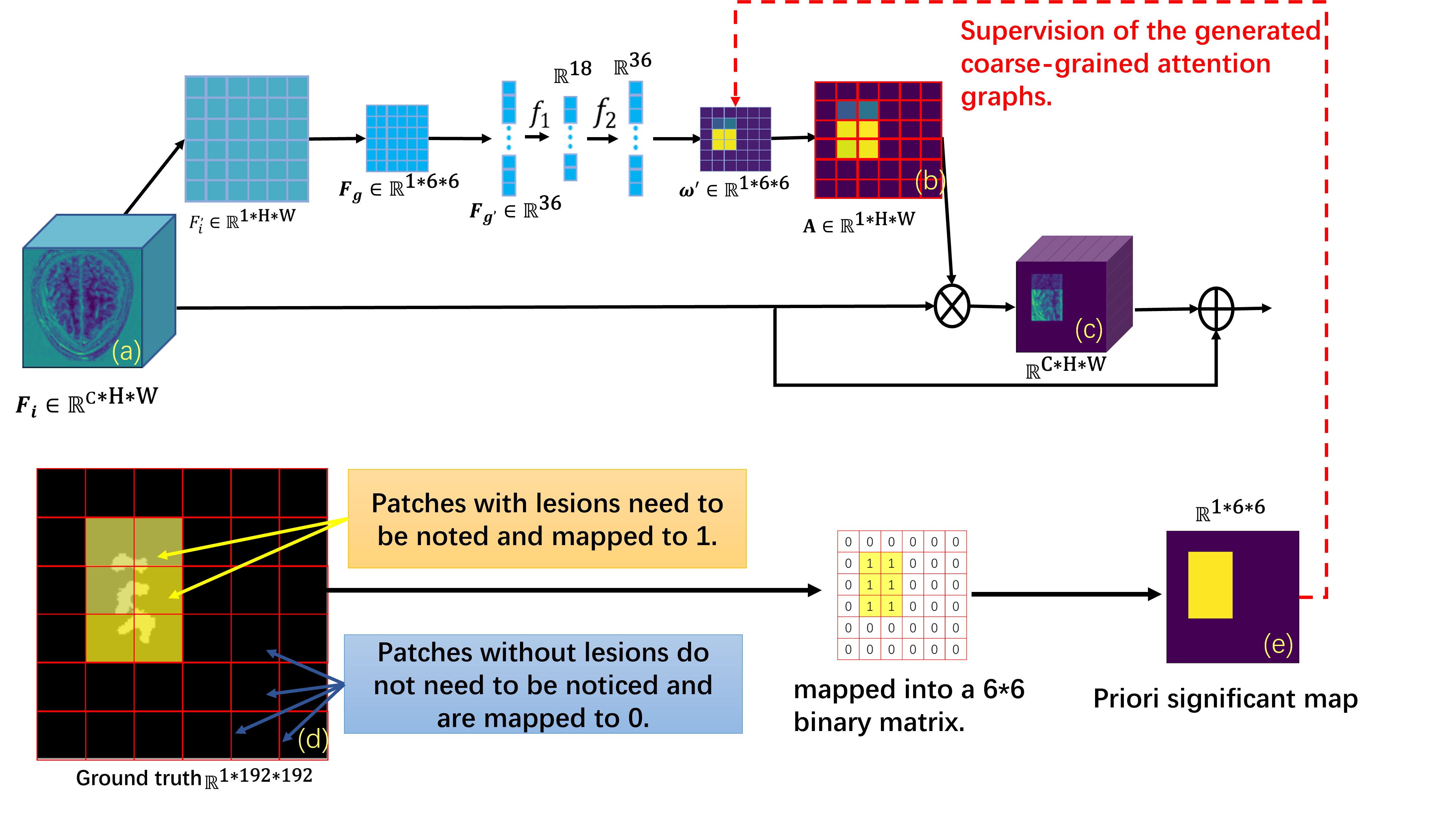} 
    \caption{The architecture of the proposed coarse-grained patch attention. Each feature map is divided into 36 equal square patches, and after the attention map, the yellow highlighted areas represent patches with a high probability of having lesions, and the darker areas represent patches with a small probability of having lesions.
}
    \label{fig:3}
\end{figure*}

As shown in Fig. \ref{fig:3}, the feature maps (Fig. \ref{fig:3}.(a)) generated from the convolution of each phase of 2D encoding in the network are used as input to the module to generate a coarse-grained attention map (Fig. \ref{fig:3}.(b)) of the same resolution size. The attention maps describe the probability of the presence of targets in different patches in the feature maps, revealing which patches to emphasize or suppress in the feature representation. As shown in Fig. \ref{fig:3}.(b and c), attention is patch-level attention, and to obtain the spatial contextual relationships among all patches, we use the method in Squeeze-and-Excitation to compress the global spatial information into a patch descriptor, as shown in the following operation:

\begin{equation}\label{eq:0}
F_i^,(1,x_i,y_j)=\frac{\sum_{\gamma=1}^{C}F_i(\gamma,,x_i,y_j)}{C}\ \ i\in[1,H]
\end{equation}
\begin{equation}\label{eq:1}
	\begin{split}
	F_g(1,x_i,y_j)&=\frac{\sum_{\gamma=1}^{h}\sum_{\delta=1}^{w}{F_i^,(1,x_{(i-1)*h+\gamma},y_{(j-1)*w+\delta})}}{hw} \\
	&i\in[1,6]\ h=\frac{H}{6}\ w=\frac{W}{6}
	\end{split}
\end{equation}

$F_i\in\mathbb{R}^{C\ast H\ast W}$ (C: number of channels, H: Height, W: Width) represents the feature maps outputted by two layers of convolution at each encoding stage. We first obtain the feature map $F_i^,\in\mathbb{R}^{1\ast H\ast W}$ using averaging pooling applied along the channel axis, which is shown to be effective in highlighting information regions \cite{30}, as shown in Eq \ref{eq:0}. Then, in order to aggregate the spatial information to obtain the global information of each patch in the feature map \cite{30}, we generate $F_g\in\mathbb{R}^{1\ast6\ast6}$ using average pooling with a pooling pool size of $\frac{H}{6}$, as shown in
 Eq \ref{eq:1}. Each pooling step covers one patch and generates a patch descriptor.


To capture the inter-patch dependencies using the patch descriptors, each patch is adaptively recalibrated to the degree of phase response. We use multilayer perceptron to explicitly model the dependencies between patches, as in Eq \ref{eq:2}. $f_1,{\ f}_2$ denote the two-layer perceptron, and $\varphi$ denotes the Sigmoid activation function. Therefore, $F_g$ is expanded into $ F_g^\prime\in\mathbb{R}^{36}$ and fed into the multilayer perceptron with 18 neurons in the middle layer and 36 neurons in the last layer, and $\omega$ is obtained by significantly mapping the sigmoid with weights of continuous values between 0 and 1. The output of each neuron represents the probability of having lesion regions for each patch block, i.e., the degree of attention required by the network.

\begin{equation}\label{eq:2}
\omega={\varphi(f}_2{(f}_1\left(F_g^\prime\right)))
\end{equation}

\begin{figure}[!h]
    \centering
    \includegraphics[width=0.45\textwidth]{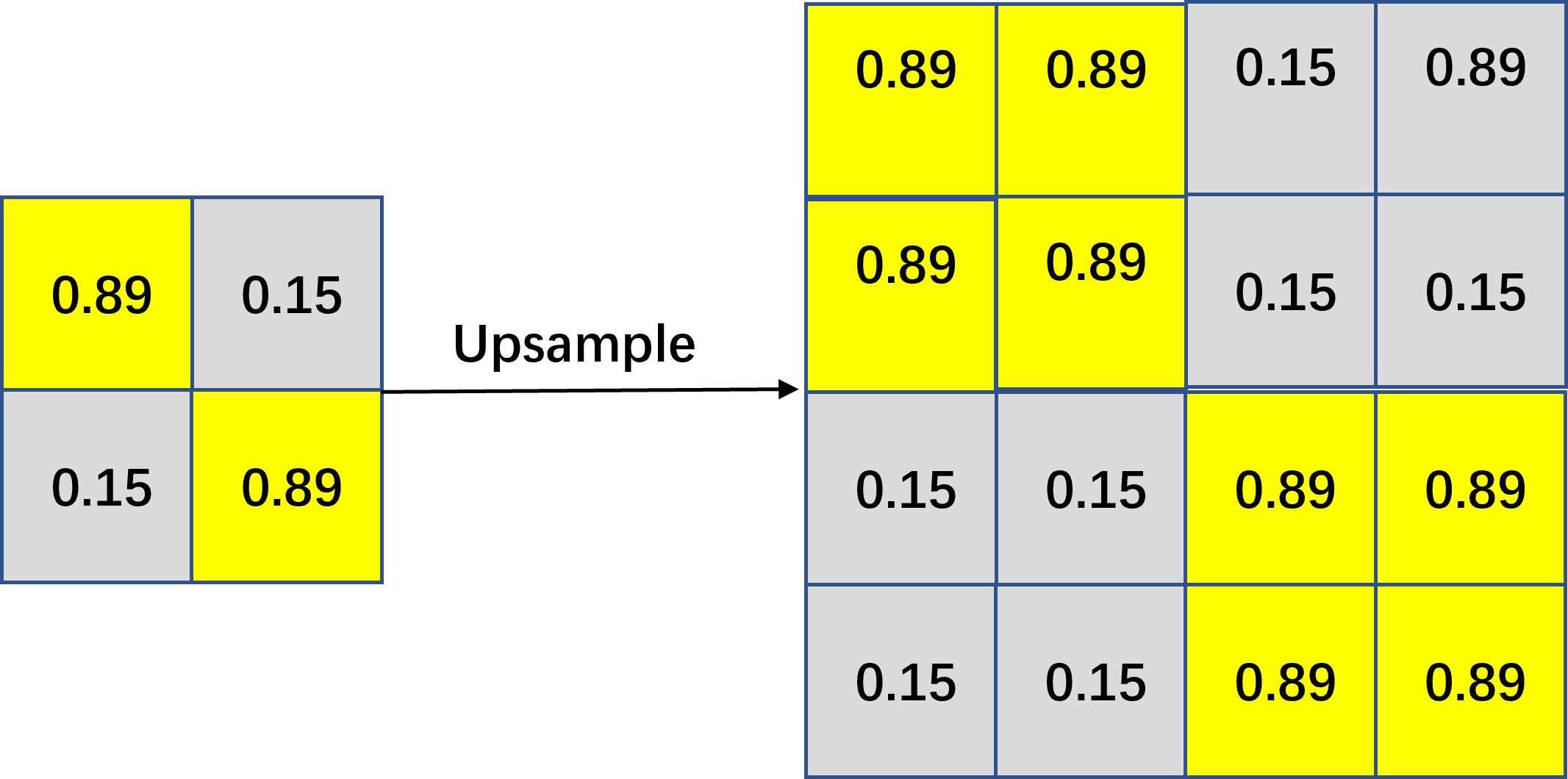} 
    \caption{The architecture of the proposed coarse-grained patch attention. Each feature map is divided into 36 equal square patches, and after the attention map, the yellow highlighted areas represent patches with a high probability of having lesions and the darker
areas represent patches with a small probability of having lesions.
}
    \label{fig:4}
\end{figure}

The obtained $\omega$ reshape into a 1*6*6 probability map $\omega^\prime$, and finally recovered into a 1*H*W patch-base coarse-grained attention map A by linear interpolation operation, where the up-sampling operation is described as Eq \ref{eq:3}, and multiplying the input feature map element-by-element level by the element, as shown in Fig. \ref{fig:4}, we can see that in the upsampled attention map (4*4), the weight scores of all pixel points in each patch (2*2) are the same, and unlike the weight scores in the fine-grained attention map are based on each pixel point, we are based on the weight scores of patches. The final output of each stage of the encoding is given by $Y=\omega_i\ast F_i+F_i$. This way provides a coarse-grained focus on the location of lesions, so that the network focuses on patches with high weight scores, and improves the target feature representation by making the features more focused on features in the target region. Note that finally, we use the idea of a residual network to add the post-attentional feature maps to the input feature maps to reduce the learning difficulty of the attentional maps while improving the error tolerance of the attentional maps.

\begin{equation}\label{eq:3}
	\begin{split}
	A(x\ast h+m,y\ast w+n)=\ \omega^\prime(x,y) \\
    x,y\in[1,6]\ h=\frac{H}{6}\ w=\frac{W}{6}
	\end{split}
\end{equation}

The CPA module learns in a supervised manner, and we generate a 1*6*6 binary attention matrix using the maximum pooling on the channel, as shown in Fig. \ref{fig:3}.(d). The size of each pooling pool is 32*32, and the patches covered by each pooling pool on ground truth are mapped as 1 whenever there is a lesion region indicating that the patches need attention; if there is no lesion, it is mapped as 0. As shown in Fig. \ref{fig:3}.(e), the binary attention matrix is visualized as a priori significant map, and we use the priori significant map as prior knowledge with binary cross-entropy loss as the coarse-grained loss function to supervise the generation of coarse-grained attention graphs in the network to obtain more accurate prediction results. More importantly, as shown in 
Fig. \ref{fig:1}.(b), although the target features become abstract with the pooling operation, the relative position of the target in the feature graph does not change. Therefore, we obtain a priori information by pooling and can supervise the generation of feature maps at each stage of encoding to ensure the accuracy of feature maps. We will also visualize the effects of attention in ablation experiments (Section \ref{sec:AblationCPA}) in CPA.

\subsection{Cross-dimensional Feature Fusion (CFF)}

The ability to extract object-specific features from CPA help networks is still limited by 2D convolution. A Cross-dimensional Fusion module of 2D features and 3D features is designed to realize the adaptive fusion of two dimensional features, replacing the naive direct addition and concatenation, the boundary feature representation accuracy of the 2D feature map is improved.

\begin{figure*}[h]
\centering
\includegraphics[width=0.9\textwidth]{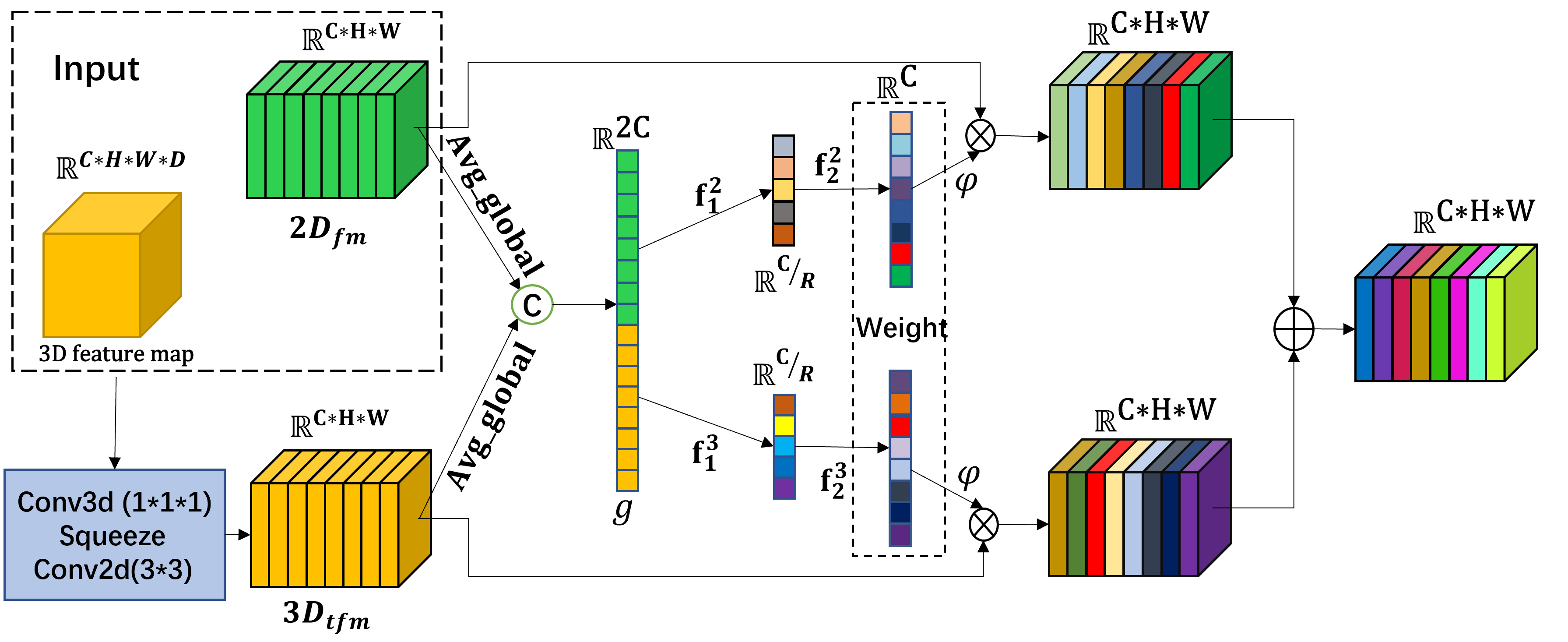}
\caption{The architecture of the proposed cross-dimensional fusion. 2D feature maps with coarse-grained patch attention and transposed 3D feature maps are used as input, and each channel feature map in different dimensions is filtered and mapped before being summed and fused.}
\label{fig:5}
\end{figure*}

The module takes 2D feature map ${2D}_{fm}\in\mathbb{R}^{C\ast H\ast W}$ and 3D feature map $\mathbb{R}^{C\ast H\ast W\ast D}$ as input (C: number of channels, H: Height, W: width, D: Depth). In each CFF module, we first downscale the 3D feature maps, then concatenate the feature maps between different dimensions and model the correlation between the combined feature channels to enhance the sensitivity of the network to features between different dimensions.

To be more specific, we first compress $\mathbb{R}^{C\ast H\ast W\ast D}$ to $\mathbb{R}^{1\ast H\ast W\ast D}$ using a single channel 1*1*1 3D convolution. Then, we use 3*3 2D convolution( filter number is set to c) to convert the 3D feature map $\mathbb{R}^{1\ast H\ast W}$ after squeezing to the 3D transformed feature map ${3D}_{tfm}\in\mathbb{R}^{C\ast H\ast W}$. Then, the 2D feature map and the 3D transformed feature map are then fed into the subsequent module, as shown in Fig. \ref{fig:5}:

From Fig. \ref{fig:5}, we can see that global average pooling is used to compress the global information of each channel of  ${3D}_{tfm}$ and ${2D}_{fm}$ into the channel descriptors, and capture the inter-channel dependencies through multilayer perceptron. We use the global information on different dimensions in series to get $g\in\mathbb{R}^{2C}$, as Eq \ref{eq:4}.  $g$ is fed into the multilayer perceptron to model the correlation of all feature channels across different dimensions. In order to reduce the computational effort, the number of neurons in the middle layer is C/R, R denotes the decay rate, and the number of neurons in the last layer is C. The channel weights $\omega_2,\omega_3\in\mathbb{R}^C$ in each dimension are obtained after significant mapping of the Sigmoid activation function, as Eq \ref{eq:5} and Eq \ref{eq:6}. $f_1^2,\ f_2^2,\ f_1^3,\ f_2^3$ denote the two-layer perceptron, $\varphi$ denotes the Sigmoid activation function, © denotes concatenate. All the channels of the two dimensions are contained in $g$, so what the multilayer perceptron learns is the relationship between all the channels between even dimensions. Our CFF applies contextual information of features in different dimensions, and if the weight score of another corresponding channel is high at the time of fusion, the deep features of that channel should be emphasized, which mutually emphasizes important features in different dimensions at the time of fusion and suppresses irrelevant features, thus enhancing the contextual semantic dependence between channel feature maps and finally enhancing the discriminative power of feature maps at the time of fusion.

\begin{equation}
\label{eq:4}
g=Avg\_global\left({3D}_{tfm}\right)\copyright Avg\_global\left({2D}_{fm}\right)
\end{equation}
\begin{equation}
\label{eq:5}
\omega_2=\varphi(f_2^2(f_1^2(g)))  
\end{equation}
\begin{equation}
\label{eq:6}
\omega_3=\varphi(f_2^3(f_1^3(g))) 
\end{equation}

The obtained weights $\omega$ are used to remap ${3D}_{tfm}$ and ${2D}_{fm}$ to obtain the feature map $output\in\mathbb{R}^{C\ast H\ast W}$ with fused 2D and 3D information, as Eq \ref{eq:7}. 
\begin{equation}
\label{eq:7}
output={3D}_{tfm}\ast\omega_3+{2D}_{fm}\ast\omega_2
\end{equation}

\subsection{Multi-scale Deconvolution Upsampling (MDU)}

In U-shaped networks, parameterless linear interpolation and simple deconvolution do not recover well the spatial and detail information lost by successive pooling in the encoding phase, making segmentation challenging. To address this problem, we propose a multi-scale deconvolution upsampling module that uses multi-scale contextual information to recover lesion region features.

\begin{figure*}[!t]
\centering
\includegraphics[width=0.95\textwidth]{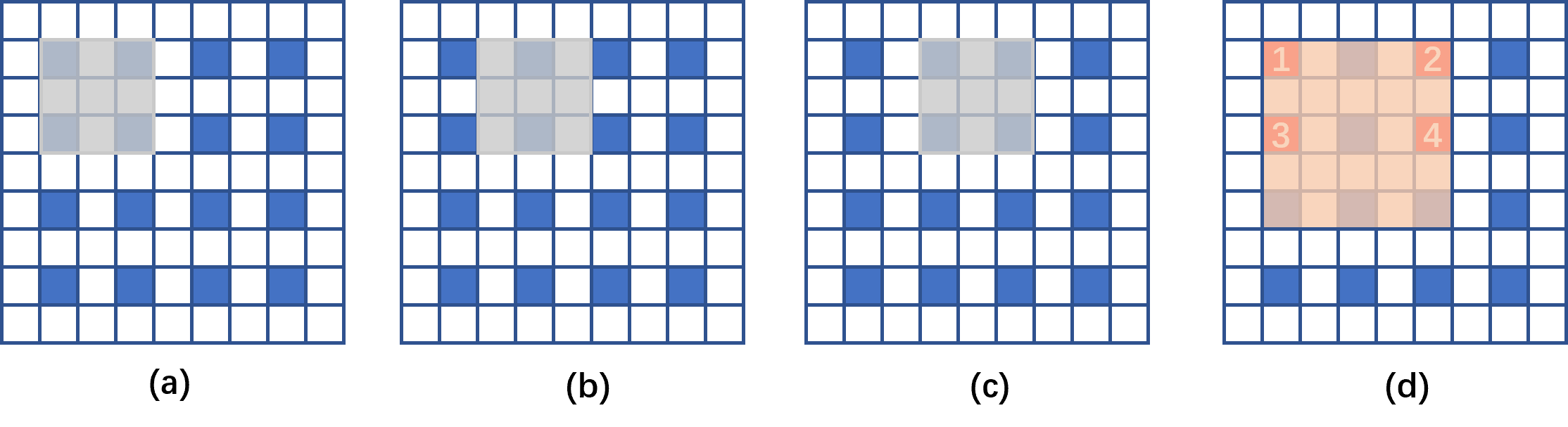}
\caption{(a, b, c): The 3*3 deconvolution moving process. (d): The coverage of the missed information by the 5*5 convolution on the 3*3 convolution.}
\label{fig:6}
\end{figure*}

As shown in Fig. \ref{fig:6}, the 9*9 white and blue feature maps are the expanded feature maps during the deconvolution process, the white area is the 0-value expanded by the deconvolution operation, the blue region is the valid region of the original input feature map, and the gray rectangle representing the 3*3 convolution operation. In the 3*3 convolution, less than half of the valid information is undoubtedly detrimental to the recovery of detailed features in the lesion region, and this problem is fatal for small targets. By comparing the three figures (a),(b) and (c) we can find that during the convolution transformation of 3*3 size, the relationship between some pixel points is ignored due to the limitation of sensory field, for example, the relationship between red pixel points 1, 2, 3 and 4 in the (d) figure is ignored. However, these pixel points can be covered by the 5*5 convolution. Similarly, 5*5 convolution has this drawback, but 7*7 convolution can make up for it. So we use multiple convolution kernels with different sizes of deconvolution to compensate for these drawbacks, which can improve the overall structure of the region by more multi-scale contextual and better recover the lesion region features. Considering the computational effort of the downsampling stage and the size of the feature map, we only use 3*3, 5*5, 7*7, and 9*9 sizes convolution.

As in Fig. \ref{fig:2}, the module $input\in\mathbb{R}^{C\ast h\ast w}$, we use a different convolution kernel size of the inverse convolution to enlarge the feature map, specifically the convolution kernel size is 3*3, 5*5, 7*7, 9*9 convolution, and the padding size is 1, 2, 3, 4, respectively, all yield an eigenmap of $\mathbb{R}^{\frac{C^\prime}{4}\ast2H\ast2W}$. Then use to concatenate them together as $output\in\mathbb{R}^{C^\prime\ast2H\ast2W}$, each convolution is followed by BN, Relu, and Dropout operations. This not only avoids overfitting and GPU memory usage but also increases the richness of the learned representations, as demonstrated by similar designs.

\section{Implementation}
\subsection{ Datasets and Preprocessing}
We evaluate our method on the open-source dataset anatomical tracing of post-stroke lesions  ATLAS  \cite{21}, which contains T1-weighted MRI scans of 229 stroke patients. The 3D image of each case contains 189 slices of 233*197. To accommodate the needs of the network, we took the 3D images of each case and sliced them along the Z-axis, and cropped the images to 180*180. Next, the image is resized to 194*194, and the first two adjacent images of each slice are selected images, and the next adjacent slice of each slice, for a total of 4 channels, as the input to each network, with no data enhancement.

To further demonstrate the effectiveness of our method, we needed image data with better 3D resolution in all three coordinate directions to fit our network structure, and we used the entire dataset from the prestigious Ischemic Stroke Lesion Segmentation Challenge (ISLES 2015) dataset \cite{56} to segment the entire tumor region as an auxiliary validation dataset, and used the same T1 modality as the input to our network. The data were preprocessed in the same way as the ATLAS data were preprocessed.

\begin{table*}[t]
\caption{The proposed architecture of AGMR-Net, where KS denotes the convolutional kernel size, CN denotes the number of channels, and Concat denotes the Concatenate operation, which connects the underlying and high-level feature maps. }
\small
\resizebox{\linewidth}{!}{
\begin{tabular}{ccccccccc}
\hline
\multicolumn{9}{c}{Attention   Guided Multiscale Recovery framework}                                                                                                                                                                                                                 \\ \hline
\multicolumn{6}{c|}{Encoding phase}                                                                                                                                                     & \multicolumn{3}{c}{Decoding phase}                                                         \\ \hline
\multicolumn{3}{c|}{3D Encoders}                                                           & \multicolumn{3}{c|}{2D Encoders}                                                           & \multicolumn{3}{c}{2D Dncoders}                                                            \\ \hline
\multicolumn{1}{c|}{Layer} & \multicolumn{1}{c|}{KS CN} & \multicolumn{1}{c|}{Output size} & \multicolumn{1}{c|}{Layer} & \multicolumn{1}{c|}{KS CN} & \multicolumn{1}{c|}{Output size} & \multicolumn{1}{c|}{layer}  & \multicolumn{1}{c|}{KS CN} & \multicolumn{1}{c}{Output size} \\ \hline
\multicolumn{3}{c|}{Input:1*4*192*192}                                                     & \multicolumn{3}{c|}{Input:192*192*4}                                                       & \multicolumn{1}{c|}{Conv2D} & \multicolumn{1}{c|}{3*3 1} & \multicolumn{1}{c}{192*192}     \\ \hline
Conv3D                     & 3*3*3 32                   & \multicolumn{1}{c|}{4*192*192}   & Conv2D                     & 3*3 32                     & \multicolumn{1}{c|}{192*192}     & Conv2D                      & 3*3 32                     & 192*192                          \\ \hline
Conv3D                     & 3*3*3 32                   & \multicolumn{1}{c|}{4*192*192}   & Conv2D                     & 3*3 32                     & \multicolumn{1}{c|}{192*192}     & Conv2D                      & 3*3 32                     & 192*192                          \\ \hline
                           &                            & \multicolumn{1}{c|}{}            & CPA                        &                            & \multicolumn{1}{c|}{192*192}     & Concat                      &                            &                                  \\ \hline
Pool                       &                            & \multicolumn{1}{c|}{2*96*96}     & Pool                       &                            & \multicolumn{1}{c|}{96*96}       & MDU                         &                            & 192*192                          \\ \hline
Conv3D                     & 3*3*3 64                   & \multicolumn{1}{c|}{2*96*96}     & Conv2D                     & 3*3 64                     & \multicolumn{1}{c|}{96*96}       & Conv2D                      & 3*3 64                     & 96*96                            \\ \hline
Conv3D                     & 3*3*3 64                   & \multicolumn{1}{c|}{2*96*96}     & Conv2D                     & 3*3 64                     & \multicolumn{1}{c|}{96*96}       & Conv2D                      & 3*3 64                     & 96*96                            \\ \hline
                           &                            & \multicolumn{1}{c|}{}            & CPA                         &                            & \multicolumn{1}{c|}{96*96}       &                             &                            &                                  \\ \hline
                           &                            & \multicolumn{1}{c|}{}            & CFF                        &                            & \multicolumn{1}{c|}{96*96}       & concat                      &                            & 96*96                            \\ \hline
Pool                       &                            & \multicolumn{1}{c|}{1*48*48}     & Pool                       &                            & \multicolumn{1}{c|}{48*48}       & MDU                         &                            & 96*96                            \\ \hline
Conv3D                     & 3*3*3 128                  & \multicolumn{1}{c|}{1*48*48}     & Conv2D                     & 3*3 128                    & \multicolumn{1}{c|}{48*48}       & Conv2D                      & 3*3 128                    & 48*48                            \\ \hline
Conv3D                     & 3*3*3 128                  & \multicolumn{1}{c|}{1*48*48}     & Conv2D                     & 3*3 128                    & \multicolumn{1}{c|}{48*48}       & Conv2D                      & 3*3 128                    & 48*48                            \\ \hline
                           &                            & \multicolumn{1}{c|}{}            & CPA                        &                            & \multicolumn{1}{c|}{48*48}       &                             &                            &                                  \\ \hline
                           &                            & \multicolumn{1}{c|}{}            & CFF                        &                            & \multicolumn{1}{c|}{48*48}       & Concat                      &                            & 48*48                            \\ \hline
                           &                            & \multicolumn{1}{c|}{}            & Pool                       &                            & \multicolumn{1}{c|}{24*24}       & MDU &                            & 48*48                            \\ \hline
                           &                            & \multicolumn{1}{c|}{}            & Conv2D                     & 3*3 256                    & \multicolumn{1}{c|}{24*24}       & Conv2D                      & 3*3 256                    & 24*24                            \\ \hline
                           &                            & \multicolumn{1}{c|}{}            & Conv2D                     & 3*3 256                    & \multicolumn{1}{c|}{24*24}       & Conv2D                      & 3*3 256                    & 24*24                            \\ \hline
                           &                            & \multicolumn{1}{c|}{}            & CPA                        &                            & \multicolumn{1}{c|}{24*24}       & Concat                      &                            &                                  \\ \hline
                           &                            & \multicolumn{1}{c|}{}            & Pool                       &                            & \multicolumn{1}{c|}{12}          & MDU &                            & 24*24                            \\ \hline
                           &                            & \multicolumn{1}{c|}{}            & Conv2D                     & 3*3 512                    & \multicolumn{1}{c|}{12*12}       &                             &                            &                                  \\ \hline
                           &                            & \multicolumn{1}{c|}{}            & Conv2D                     & 3*3 512                    & \multicolumn{1}{c|}{12*12}       & \textbf{}                   &                            &                                  \\ \hline
                           &                            & \multicolumn{1}{c|}{}            & CPA                        &                            & \multicolumn{1}{c|}{12*12}       &                             &                            &                                  \\ \hline
\end{tabular}
}
\label{tab:2}
\end{table*}

\subsection{ Implementation Details}
Our proposed AGMR-Net is initialized using the Kaiming Initialization method. The detailed configuration is shown in Table \ref{tab:2}, with a given input data of 192*192*4. The size of the 2D convolutional kernel in the encoding stage is 3*3, the size of the 3D convolutional kernel is 3*3, and the size of the 2D convolutional kernel in the decoding stage is 3*3. concatenate connects the features in the encoding and decoding stages. Each convolutional layer is followed by a normalization layer and a ReLU activation layer. The optimizer used to train the network in the training phase is Adam, with an initial learning rate of 10-3. Also, for the stability of the network training, our learning rate decay strategy uses exponential decay, with a decay floor of 0.96. The deep learning framework we use is PyTorch, version Linux 1.2.0. Each layer uses a different Dropout to prevent overfitting of the network and is trained for 150 epochs before the network is tested. The tests for the training of the experiments were run on NVIDIA Tesla V100S GPUs.

\subsection{Quantitative Evaluation}
To quantitatively compare segmentation results on the test set, we used six segmentation metrics: (1) the Dice similarity coefficient (DSC), (2) DSC (global) represents the metric based on the voxel calculation, (3) recall, (4) precision, (5) the Average symmetric Surface Distance (ASD) (6) the 95 percent Hausdorff Distance (HD). DSC is used to estimate the spatial overlap between prediction and ground truth, HD is an anomaly robust metric based on the Hausdorff distance between the two boundaries, ASD is the average of all distances from points on the predicted surface to ground truth, and DSC (global) provides a more visual representation of the overall DSC score based on voxels. Where the metrics of DSC, recall, precision, ASD, and HD are based on the mean ± standard deviation of each patient in the test set. The number of true-positive, false-positive, true-negative, and false-negative test samples, expressed as TP, FP, TN, and FN, were counted by the final segmentation results and ground truth. Then we have:

\begin{equation}
Precision=\frac{TP}{TP+FP}  
\end{equation}
\begin{equation}
Recall=\ \frac{TP}{TP+FN}  
\end{equation}
\begin{equation}
DSC=\frac{2TP}{2TP+FP+FN} 
\end{equation}

\subsection{ Loss function}
To explicitly refine the features as well as the segmentation results, we use augmented mixture loss \cite{31} in the network training to supervise the final segmentation results, as in Eq \ref{eq:11}, consisting of Focal loss \cite{53} and Dice loss \cite{54}. On top of this, we use the Sobel loss \cite{55} function to better refine the bounds of the segmentation results and the cross-entropy loss function to supervise the attention map of the coarse-grained patch attention output for each layer. To prevent excessive influence on the final segmentation loss, we set the percentage of coarse-grained attention loss to 0.15. Our final loss function consists of three components. As in Eqs  \ref{eq:11}--\ref{eq:13}, where p is the result predicted by our method and g is the ground truth. $cp$ denotes the coarse-grained prior information and $cg_i$ denotes the coarse-grained attention graph generated at each stage.

\begin{equation}
L_1=\frac{1}{N}Foccal\_loss(p,g)-log{\left(Dice\_loss(p,g\right))} 
\label{eq:11}
\end{equation}
\begin{equation}
L_2=Sobel\_loss(p,g) 
\label{eq:12}
\end{equation}
\begin{equation}
L_{3-i}=BCE\_loss(cp,cg_i) i\in{1,2,3,4,5} 
\label{eq:13}
\end{equation}
\begin{equation}L=L_1+L_2+0.15\sum_{i=1}^{5}L_{3-i}
\label{eq:14}
\end{equation}

\section{Results and Discussion}
\subsection{Comparison with State-of-the-Art Methods on ATLAS}

The AGMR-Net has been validated by comparison with two advanced fine-grained spatial attention methods Attention Unet \cite{34}, and the self-attention based method X-net \cite{5} , as well as the 2D and 3D fusion network D-Unet \cite{31}, and three other more advanced multi-scale network methods Pspnet \cite{17}, Deeplab V3 \cite{16}, and CLCI-Net \cite{4} methods. The advanced networks Segnet \cite{58}, U-Net \cite{12}, 3D-Unet \cite{59} and Unet++  \cite{29} for medical image segmentation have been performed. The quantitative analysis of segmentation is shown in Table \ref{tab:3}. Our AGMR-Net outperforms eleven state-of-the-art algorithms on the ATLAS stroke segmentation datasets, segmenting liver tumors with DSC of 0.594, DSC(global) of 0.754, Recall of 0.579, and Precision of 0.713. CPA, CFF, and MDU are responsible for these high segmentation and detection results.

\begin{table*}[h]
\caption{ Comparison of the accuracy of our proposed method with other well-known methods on the Atlas dataset, "-" indicates that the corresponding result is not given or cannot be measured}
\resizebox{\linewidth}{!}{
\begin{tabular}{ccccc}
\hline
Method         & DSC                  & DSC(global)    & Recall      & Precision            \\ \hline
3D   Unet      & 0.471±0.287          & 0.709          & 0.474±0.310 & 0.553±0.325          \\
Unet           & 0.485±0.298          & 0.712          & 0.469±0.296 & 0.574±0.328          \\
Segnet         & 0.451±0.293          & 0.6629         & 0.440±0.304 & 0.545±0.301          \\
Pspnet         & 0.468±0.283          & 0.6723         & 0.443±0.284 & 0.539±0.307          \\
Deeplab v3     & 0.415±0.263          & 0.6411         & 0.399±0.266 & 0.492±0.299          \\
Attention Unet & 0.536±0.261          & 0.706          & 0.482±0.272 & 0.721±0.254          \\
Unet++         & 0.511±0.291          & 0.728          & 0.494±0.484 & 0.638±0.352          \\
D-UNet         & 0.548±0.264          & 0.7269         & 0.530±0.268 & 0.665±0.303          \\
X-net          & 0.499±0.300          & 0.6724         & 0.466±0.303 & 0.633±0.312          \\
CLCI-Net       & 0.581                & -              &\textbf{ 0.581}       & 0.649                \\
Ours           & \textbf{0.594±0.273} & \textbf{0.754} & 0.579±0.287 & \textbf{0.713±0.275} \\ \hline
\end{tabular}}
\label{tab:3}
\end{table*}

\begin{figure*}[h]
\centering
\includegraphics[width=0.95\textwidth]{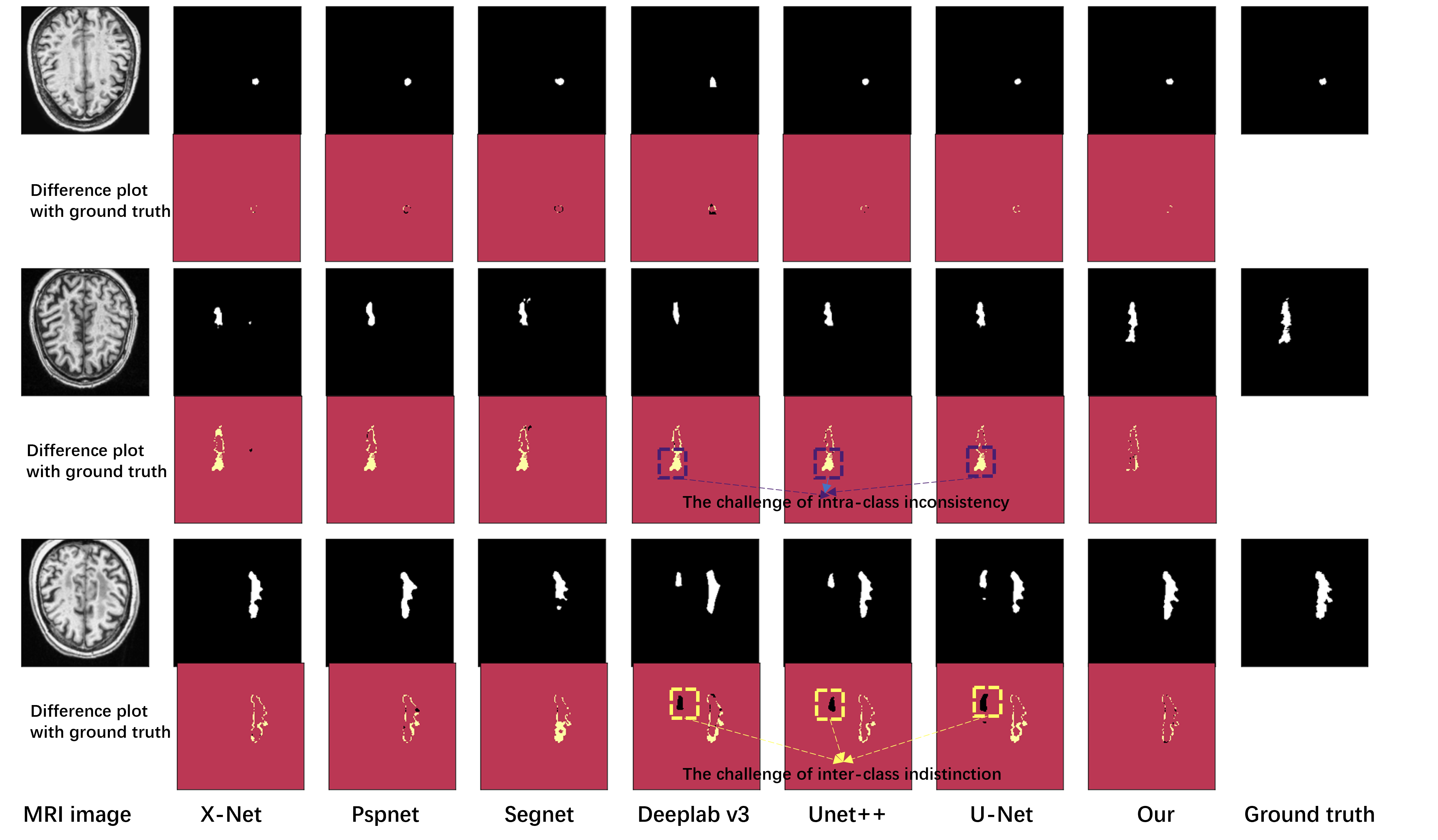}
\caption{Segmentation results of one of the method pairs when facing intra-class inconsistency and inter-class indistinction.}
\label{fig:7}
\end{figure*}

We make the difference between the prediction results of the eight methods and the ground truth, as shown in Fig.  \ref{fig:7}, to more conveniently display the effect of our methods on intra-class inconsistency and inter-class indistinction difficulties. Due to intra-class inconsistency, the network predicts the wrong lesion (false negative) in the yellow area of the difference plot, and due to inter-class indistinction, the network predicts the wrong normal tissue (false positive) in the black area. As can be seen, our technique produces the fewest false-negative and false-positive regions, effectively improving the network's segmentation performance.

\begin{figure*}[h]
\centering
\includegraphics[width=0.95\textwidth]{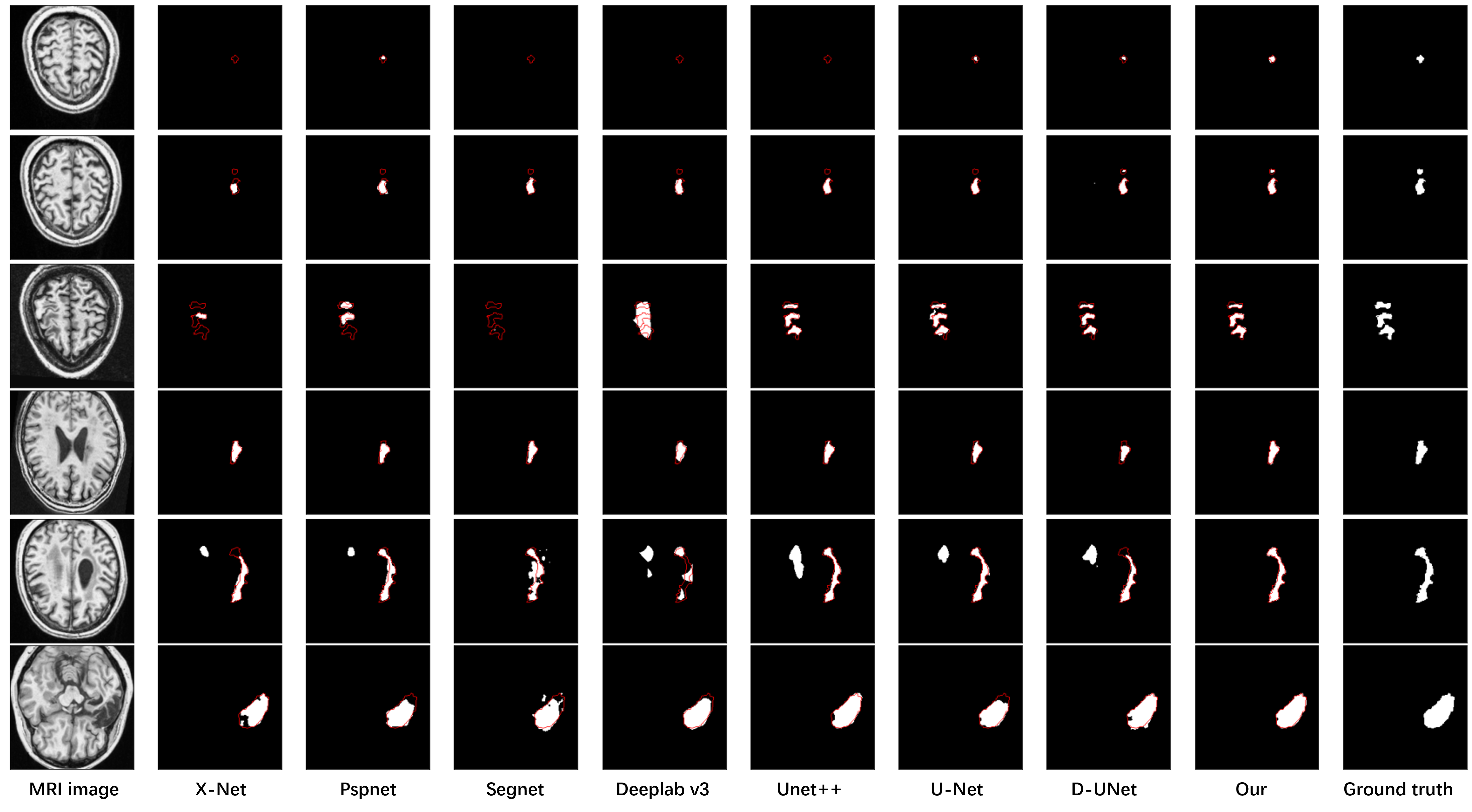}
\caption{Comparisons of our method, X-net, Pspnet, Segnet, Deeplab v3, Unet++, and U-Net on five different patients.}
\label{fig:8}
\end{figure*}

We also visually compare the stroke boundary segmentation results of our method with other methods. As shown 
in Fig. \ref{fig:8}, the red line in the figure is the boundary of ground truth, and since CFF produces more accurate boundary features in the encoding stage and MDU recovers the boundary features well in the encoding stage, we can see that our segmentation results are closest to the true boundary. Unet++ and CLCI-Net, although they modify the skip connection in the encoding stage to achieve better recovery of the target The effect of boundary, Deeplab V3 also uses multi-scale information, but it is still insufficient in the boundary segmentation effect. Table \ref{tab:4} also shows quantitatively how our method compares with other methods on boundary segmentation.

We obtained an ASD of 7.137 and a 95HD of 27.005. And we make the difference between the obtained ASD and HD with the results of other networks, as shown in Table \ref{tab:4}, the ASD is 1.795 points lower and the 95HD is 9.550 points lower than our baseline D-Unet. The reduction is also compared to other networks, which proves that our method obtains a better boundary segmentation effect. We used DSC scores to run a significance analysis between our model and other traditional models to see if there was a meaningful improvement, as shown in Table \ref{tab:4}. All of the p-values after the experiments were less than 0.05, and p-values less than 0.05 were considered significant, demonstrating the effectiveness of our strategy in comparison to other models. In addition, when compared to the baseline D-Unet, our method's P-value is substantially smaller than 0.05.

\begin{table}[!h]
\centering
\caption{ P-values in the significance analysis were compared. All P-values are less than 0.05, which proves that the proposed model has a significant advantage.}
\begin{tabular}{llll}
\hline
Method                & ASD    & 95HD    & p-value \\ \hline
Ours-Unet             & -8.008 & -25.970 & 1.35E-5 \\
Ours-Segnet           & -5.089 & -18.669 & 1.92E-6 \\
Ours-Pspnet           & -8.269 & -17.908 & 4.34E-7 \\
Ours-Deeplab   v3     & -7.058 & -14.594 & 4.77E-7 \\
Ours-Attention   Unet & -0.811 & -14.637 & 0.002   \\
Ours-Unet++           & -6.827 & -10.542 & 4.06E-4 \\
Ours-D-UNet           & -1.795 & -9.550  & 0.001   \\
Ours-X-net            & -4.474 & -19.497 & 9.40E-5 \\ \hline
\end{tabular}
\label{tab:4}
\end{table}

\begin{figure}[!h]
\centering
\includegraphics[width=0.9\textwidth]{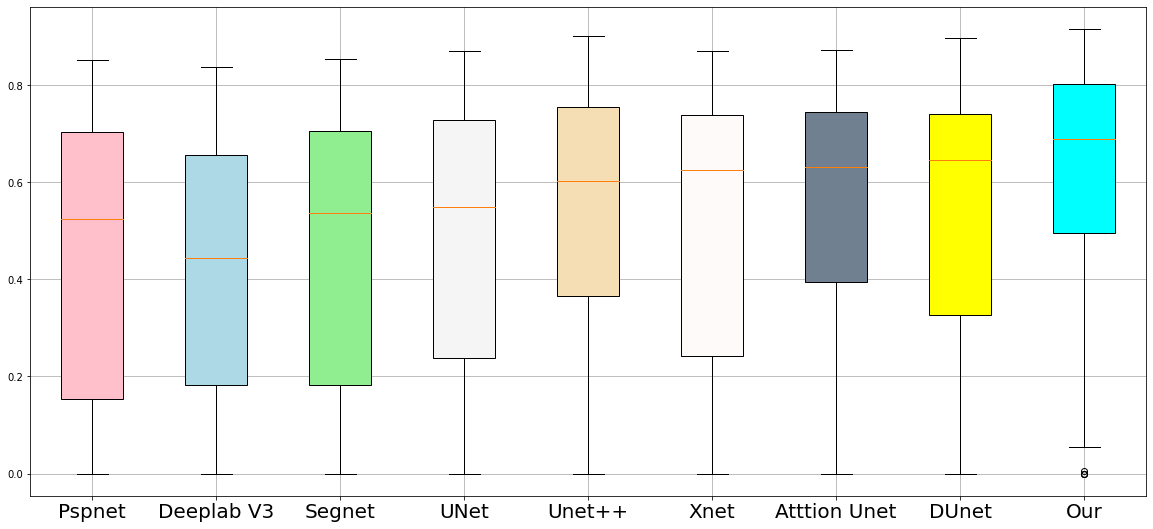}
\caption{The proposed AGMR-Net approach was compared with eight state-of-the-art models evaluated using DSC scores. The proposed model achieved the highest mean and median DSC scores.}
\label{fig:9}
\end{figure}

We also display the box plot results on the test set, as shown in Fig. \ref{fig:9}, where the DSC coefficients were utilized to evaluate the outcomes. Our model has the highest median DSC value of 0.594, suggesting that it performs well when compared to the other models, and it also has the highest DSC score, showing that it works well. The model's DSC scores were frequently focused at the higher end, with the highest quartile of all nine approaches.

\subsection{Ablation Studies of Our Method}
To analyze the impact of each module, we performed ablation experiments using these modules and their combinations, and the results of the ablation experiments are shown in Table \ref{tab:5} to verify that the design modules are effective for improving segmentation accuracy. Specifically, we use CPA, CFF, and MDU to represent the three modules respectively.

\begin{table*}[]
\centering
\caption{ Verify the impact of each module on the baseline.}
\resizebox{\linewidth}{!}{
\begin{tabular}{cccccccccc}
\hline
CPA & CFF & \multicolumn{2}{c}{MDU}       & DSC(global)      & DSC                  & Recall               & precision            & ASD                   & HD                     \\ \hline
    &     &             & \multicolumn{2}{c}{0.726}          & 0.548±0.264          & 0.530±0.268          & 0.665±0.303          & 8.932±16.565          & 36.552±30.262          \\
$\surd$   &     &             & \multicolumn{2}{c}{0.741}          & 0.574±0.272          & 0.561±0.290          & 0.682±0.297          & 7.833±13.900          & 33.544±30.346          \\
    & $\surd$   &             & \multicolumn{2}{c}{0.734}          & 0.572±0.244          & 0.538±0.244          & 0.701±0.271          & 8.670±17.513          & 30.197±29.383          \\
    &     & $\surd$           & \multicolumn{2}{c}{0.735}          & 0.570±0.276          & 0.541±0.274          & 0.701±0.302          & 7.972±15.818          & 27.220±29.333          \\
$\surd$   & $\surd$   &             & \multicolumn{2}{c}{0.744}          & 0.581±0.269          & 0.569±0.277          & 0.667±0.302          & 7.521±16.156          & 29.174±28.246          \\
$\surd$   & $\surd$   & $\surd$           & \multicolumn{2}{c}{\textbf{0.754}} & \textbf{0.594±0.273} & \textbf{0.579±0.287} & \textbf{0.713±0.275} & \textbf{7.137±14.988} & \textbf{27.005±30.353} \\ \hline
\end{tabular}}
\label{tab:5}
\end{table*}

It can be seen that adding our proposed modules all results in a higher DSC score than baseline. we propose the CPA module, which adaptively highlights the target spatial information. Although it is a coarse-grained form of salient representation that enhances the target information while also enhancing some of the surrounding background information. However, the precision and recall of our method are significantly improved, and the DSC is improved to 0.574, which indicates that CPA can guide the network to extract the features of the target in the salient region, solve the inter-class differences and suppress the background noise well. The DSC reaches 0.572 and the HD drops to 30.197, indicating that we obtain a more accurate boundary segmentation, while the accuracy improves to 0.701, effectively reducing false negatives, indicating that the CFF module can well solve the challenge of interclass variation. The combination of the two modules also improves the DSC to 0.581 and reduces both ASC and HD. However, the above two modules are limited to the encoding stage, and the spatial and boundary information highlighted by the modules are lost due to pooling operations. For this reason, the upsampling phase of our decoding stage better recovers the detailed information of the high-level semantic features by multi-scale deconvolution of the upsampling modules. With the combination of all modules, we achieved the best results, which demonstrates the effectiveness of all components.

\subsection{ Ablation Study for CPA Module}\label{sec:AblationCPA}
{\bf (1)Comparison with fine-grained attention.} To evaluate the effectiveness of our proposed CPA module, we further compare it with other location attention methods: classical network Attention-Unet, Baseline+FA (integrated with the fine-grained attention approach in CBAM). The results are shown in Table \ref{tab:6}, our CPA approach obtains the best performance in ATLAS, due to the other two fine-grained approaches, DSC and DSC(global) accuracy are higher than Baseline+FA 0.023 and 0.017.

\begin{table}[!h]
\centering
\caption{  Comparison with other positions of attention}
\begin{tabular}{ccc}
\hline
Method         & DSC         & DSC(global) \\ \hline
Attention-Unet & 0.536±0.261 & 0.706       \\
Baseline + FA  & 0.551±0.253 & 0.724       \\
Baseline + CPA & 0.574±0.272 & 0.741       \\ \hline
\end{tabular}
\label{tab:6}
\end{table}

\begin{figure}[!h]
\centering
\includegraphics[width=0.9\textwidth]{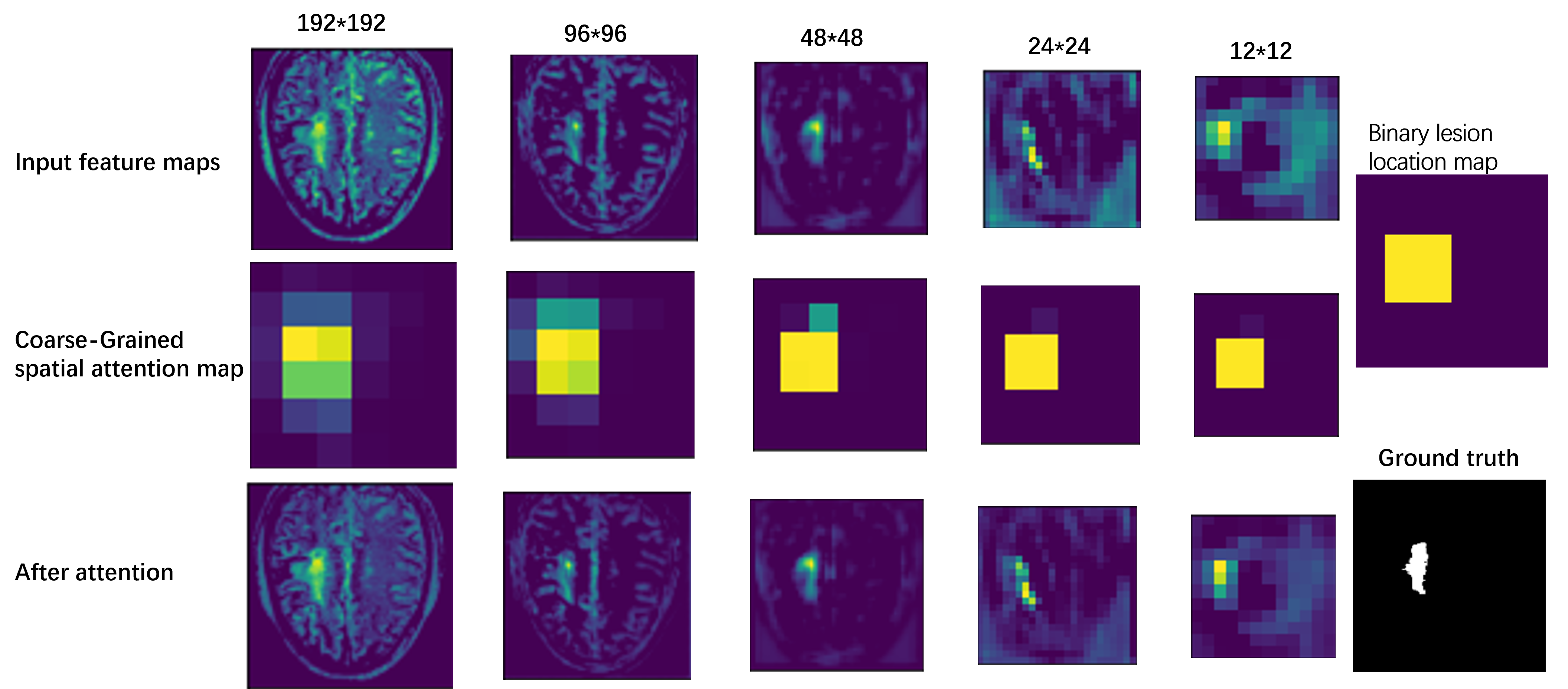}
\caption{The first to fifth columns show the feature map illustrations in the five coarse-grained location attention modules in the network coding stage. The first to the third rows are the input feature maps, coarse-grained attention maps, and output features of the module, respectively. The rightmost part of the figure shows the ground truth as well as the a priori binary lesion location map.}
\label{fig:10}
\end{figure}

{\bf (2)Visible CPA module for significant representation of target spatial information.} Here we visualize the effect of the CPA module, as shown in Fig. \ref{fig:10}. The second row shows the attention map generated by our coarse-grained patch attention, where the brightly colored regions represent the patches that the network needs to pay attention to, the target regions have a high probability of being present in these patches. Comparing the contrast between the background and lesion regions of the feature maps before and after attending, the task-relevant regions are significantly enhanced, indicating that the module has good localization ability and the ability to help the network extract features better, and try to exclude the interference of irrelevant features and reduce noise, providing a low-noise feature map during high and low-level feature fusion. At the same time, our attention map allows for effective supervision, and as the number of network layers deepens, our attention map becomes more accurate and more consistent with the a priori map generated based on ground truth, although the attention map of the first and second columns is not as accurate as the last three columns, the contrast between the patches of the target region and other patches in the attention map is still obvious, and we borrow the idea of Resnet to make the feature map of important information will not be suppressed because of the shortage of the pre-attentive map. And it can be seen that because the relative position of the target in the feature map changes little in each layer, our proposed CPA is achievable by multi-layer supervision to generate accurate attention maps, and the patched-based attention approach can accurately highlight the spatial information of the target in each layer and enhance the contrast between the target and the background.

\begin{figure}[!h]
\centering
\includegraphics[width=0.8\textwidth]{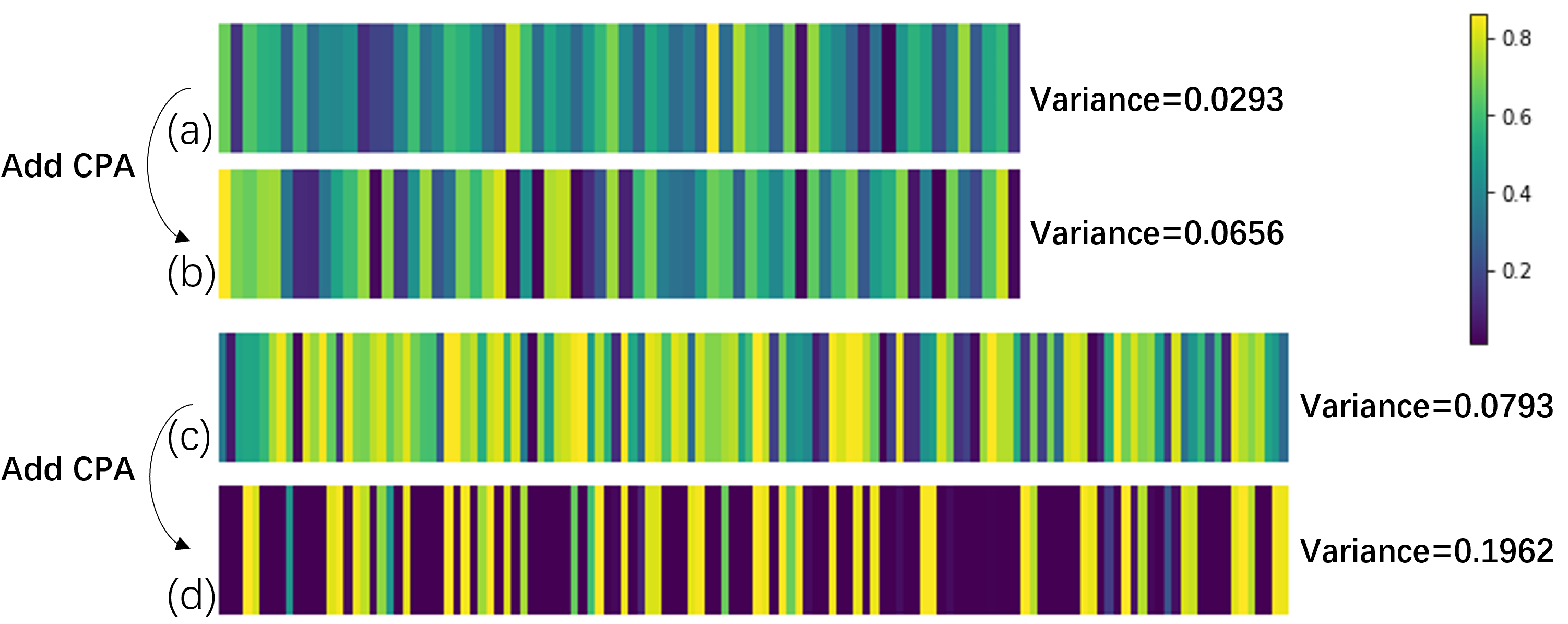}
\caption{The vector of attention values for the 2D branch weights of the two phases with the addition of CFF and with the addition of CFF and CLA are shown. The color bar on the right side is the color corresponding to the weight values.}
\label{fig:11}
\end{figure}

{\bf (3)Significant representation of target spatial information can guide the network to extract target-specific features.} To illustrate more effectively the CPA module can guide the network to extract target-related features, we quantitatively support this by the weights of each channel in the cross-dimensional feature fusion module. Because high-quality features allow better identification of segmentation lesions, the variability between each feature increases and each channel is a response to the input feature map to the feature, so the variability between channels increases, and the difference between each channel weight value increases {28}. As shown in Fig. \ref{fig:11}.(b) and (d) are the weights of all channels of the 2D feature map after adding the cross-dimensional feature fusion module, respectively. Observing the vector map, the color difference of the vector map increases after adding CPA at each stage, both the channel weight value variance increases, especially in (d), the difference of the weight value variance is more obvious, which proves that the network extracts higher quality features after the CPA module. This is because as the network deepens, our CPA module provides more accurate information about the location of the lesion, as shown in Fig. \ref{fig:11}, and the extracted features are more differentiated and related to the segmentation target, which will also cause the difference between channels to increase.

\subsection{Ablation Study for CFF Module}
In this section, we evaluate the portability of our CFF module with several different promising modules, and can have the following observations:

{\bf (1)The weighted filtering of features is beneficial for accuracy.} Two dimensional features fusion uses simple concatenate or add to introduce useless or redundant information. In contrast, by incorporating the cross-dimensional of two dimensional global information and adaptively re-weighting multidimensional output feature mapping, our CFF module could achieve more accurate prediction. As shown in Table \ref{tab:5}, the accuracy gains on DSC, DSC (global), ASD, and HD were 0.024, 0.008, 0.262, and 6.29, respectively, compared to the D-Uet benchmark, is because the methods in D-Uet do not simply model the relationships of single-dimensional features, whereas CFF captures the dependencies between all dimensional features. Finally, we use to add or concatenate instead of the CFF module in AGMR-Net, as shown in Table \ref{tab:7}, it can be seen that the weighted filtering before feature fusion is better than the direct summation or connection, which can enhance feature fusion. We also replace the CFF type add operation with concatenate (CFF + concatenate). Although there is not much difference between DSC and AGMR-Net, the cascade increases the number of characteristic channels and the convolution operation, so we propose a CFF module using ADD.

\begin{table}[!h]
\centering
\caption{Comparison with common feature fusion approaches.}
\begin{tabular}{ccc}
\hline
Method     & DSC             & DSC(global) \\ \hline
Add        & 0.588±0.248     & 0.750       \\
Concat     & 0.583±0.257     & 0.749       \\
CFF-Concat & 0.593 ±   0.257 & 0.751       \\
AGMR-Net   & 0.594±0.273     & 0.754       \\ \hline
\end{tabular}
\label{tab:7}
\end{table}

{\bf (2)The global information is effective for feature modeling and filtering.} In Table \ref{tab:8}, we found a significant improvement in our CFF fusion technique for fusing global information when compared to the baseline’s local information fusion strategy, achieving a mean DSC of 0.572 and our p-value of 0.001. To emphasize the importance of global information, we reduce the decay rate in CFF to 32 to reduce the parameters in CFF. The number of parameters in the CFF is $2C\ast\ \frac{2C}{32}\ +2(\frac{2C}{32}*C)=\ \frac{C^2}{4}$, which is the same as the number of parameters in the feature fusion module in our baseline, $2(C*\frac{C}{16} + \frac{C}{16}*C)=\ \frac{C^2}{4}$, and the improvement of the effect is still significant compared to the baseline's P-value of 0.039. This indicates that the enhancement of the CFF module is due to the global information modeling and not due to the addition of model parameters.


\begin{table}[!h]
\centering
\caption{Global information and parametric quantities for feature fusion of images.}
\begin{tabular}{ccc}
\hline
Method      & DSC(Global) & DSC         \\ \hline
Baseline    & 0.726       & 0.548±0.264 \\
+ CFF(R=32) & 0.731       & 0.566±0.261 \\
+ CFF(R=8)  & 0.734       & 0.572±0.244 \\ \hline
\end{tabular}
\label{tab:8}
\end{table}

\subsection{Ablation Study for MDU Module}
{\bf (1)Multi-scale information is more conducive to recovering target features.} As shown in Fig. \ref{fig:1}.(b), we can see that the feature map after four times of downsampling is very abstract. At this time it is still too coarse to use simple interpolation upsampling or single deconvolution methods, so it is necessary to complete the upsampling operation while recovering the lesion details better with the help of multi-scale contextual information. It can also be seen from Table \ref{tab:5} that the segmentation effect is improved by adding MDU, the increase in recall reflects the accuracy of the segmentation location of our method, and the ASD and HD also reflect the improved segmentation boundary effect. In Table \ref{tab:9}, we compare it with the single-scale deconvolution method, and the results show that the DSC score decreases as the convolution kernel size gradually increases when the convolution kernel size is 3, 5, 7, and 9. The DSC improves only when the convolutional kernel size is 9. However, using all convolutional kernels of size 9 overloads the network and ignores the diversity of lesion sizes, resulting in a lower average DSC than our multiscale approach.

\begin{table}[!h]
\centering
\caption{The ablation practice of multi-scale deconvolution, which only simply expands the perceptual field, is not conducive to the recovery of target features.}
\begin{tabular}{cccc}
\hline
Method                  & DSC         & DSC(global) & Parameters \\ \hline
Size= 3,5,7,9 & 0.594±0.273 & 0.754       & 23799283   \\
Size = 3                & 0.579±0.278 & 0.745       & 12658163   \\
Size = 5                & 0.552±0.265 & 0.727       & 18228723   \\
Size = 7                & 0.578±0.739 & 0.735       & 26584563   \\
Size = 9                & 0.587±0.251 & 0.753       & 37725683   \\ \hline
\end{tabular}
\label{tab:9}
\end{table}

{\bf (2)The recovery of the upper sampling phase target should not be ignored.} Many existing methods capture the multiscale information of the target at the encoding or skip connection, ignoring the necessity of using multiscale information at the sampling stage on the network for recovering the target space and boundary information. As shown in Table \ref{tab:10}, we replace the MDU module with the coding phase of the network and the skip connection phase of the upper and lower interconnections. Although both are improved by adding the MDU module to the encoding phase or the jump phase compared with Baseline+CPA+CFF, the segmentation results are still inferior to our network.

\begin{table}[!h]
\centering
\caption{A comparative test of MDU module integration in different locations of the network}
\begin{tabular}{ccc}
\hline
Method                  & DSC         & DSC(global) \\ \hline
Decoder   Phase         & 0.586±0.523 & 0.748       \\
Skip   connection Phase & 0.583±0.278 & 0.748       \\
AGMR-Net                & 0.594±0.273 & 0.754       \\ \hline
\end{tabular}
\label{tab:10}
\end{table}

\subsection{Results of ISLES 2015}
We repeated the experiment using ISLES 2015 data to further validate the efficacy of our method. Because our network has four input channels, we used the same modal T1 modality as ATLAS for our tests to avoid overloading the channels with too much modal data. Table \ref{tab:11} presents the conclusions. The best segmentation accuracy of 0.614 may be attained utilizing our proposed AGMR-Net model.

\begin{table*}[]
\centering
\caption{Comparison of the accuracy of our proposed method with other well-known methods on the Atlas dataset.}
\resizebox{\linewidth}{!}{
\begin{tabular}{ccccc}
\hline
Method         & DSC                  & DSC(global)    & Recall               & Precision            \\ \hline
Unet           & 0.504±0.271          & 0.658          & 0.512±0.314          & 0.601±0.265          \\
Segnet         & 0.256±0.258          & 0.410          & 0.242±0.258          & 0.315±0.272          \\
Pspnet         & 0.393±0.269          & 0.481          & 0.607±0.371          & 0.316±0.226          \\
Deeplab v3     & 0.414±0.351          & 0.633          & 0.421±0.377          & 0.642±0.382          \\
Attention Unet & 0.516±0.340          & 0.686          & 0.559±0.368          & 0.502±0.338          \\
Unet++         & 0.316±0.222          & 0.384          & 0.617±0.365          & 0.218±0.164          \\
D-UNet         & 0.595±0.287          & 0.736          & 0.593±0.308          & \textbf{0.774±0.118} \\
X-net          & 0.437±0.320          & 0.601          & 0.419±0.323          & 0.551±0.287          \\
CLCI-Net       & 0.405±0.367          & 0.643          & 0.423±0.413          & 0.454±0.367          \\
Ours           & \textbf{0.614±0.317} & \textbf{0.753} & \textbf{0.623±0.327} & 0.633±0.345          \\ \hline
\end{tabular}}
\label{tab:11}
\end{table*}

\subsection{ Discussion}
In this paper, we propose an Attention Guided Multiscale Recovery framework for the 3D stroke lesion segmentation task. segmentation experiments were performed on the stroke lesion datasets ATLAS and ISLES 2015, and by comparing Tables \ref{tab:3}, Table \ref{tab:4}, Table \ref{tab:11}, and Fig. \ref{fig:7} with other networks, we demonstrate that the enhancement is significant.

The comparison of different granularity attention in Table \ref{tab:6} demonstrates that our proposed coarse-grained attention module is fast due to the traditional fine-grained attention model. These improvements are attributed to the fact that our coarse-grained attention has explicit multilayer supervision, which can accurately highlight features useful for the stroke segmentation task and improve the segmentation capability of the network by suppressing irrelevant regions in the feature images. the capability of the CPA structure can be verified by the heat map in Fig. \ref{fig:10}. And Fig. \ref{fig:11}  demonstrates that the CPA module can guide the network to extract more effective features. Fig. \ref{fig:7} also demonstrates that the AMGR-Net with the addition of the CPA module can effectively cope with the challenge of intra-class inconsistency. However, the patched-based attention approach also limits the resolution of the input images and the number of layers of the network, too large patches also enhance a lot of background information, and too small patches are not suitable for use in the deeper layers out of the network. So in the future, we will explore the use of multi-granularity attention.

From Table \ref{tab:7} and Table \ref{tab:8}, we demonstrate that our CFF module has a significant improvement over the traditional concatenate and add, even compared to the baseline that inherits the advanced channel attention, due to the fact that we model the global information of two different tasks. In Tables \ref{tab:8}, we demonstrate that our approach does not improve performance by simply increasing the amount of computation. Fig. \ref{fig:8} also proves that the AMGR-Net after adding the CFF module gets the most bounded partitioning that helps to solve the intra-class inconsistency challenge.

Table \ref{tab:9} and Table \ref{tab:10} demonstrate that we use multiscale recovery in the upsampling phase. However, we see from Table \ref{tab:9} that our MDU module is more computationally intensive compared to the 3*3 deconvolution, which is the goal of our future improvement to reduce the computation of the upsampling module and at the same time recover the information loss caused by pooling more efficiently.

\section{Conclusions}
Our AGMR-Net integrates CPA and CFF modules in the encoding stage. The former uses a coarse-grained attention map to focus the network on the target region to extract features highly relevant to the target, and the latter fuses feature from 3D for the former to refine the boundaries of the 2D feature map. Finally, the MDU is used in the decoding stage to recover the lesioned features from the feature maps that have been downsampled several times. The main advantage of our AGMR-Net is the use of spatial contextual information between each feature and different patches, as well as global contextual information between different dimensional channels and multi-scale contextual information, showing that the application of contextual information is crucial for performance improvement and demonstrates a great advantage when compared with existing methods.

\section{Acknowledgements}
We thank all anonymous reviewers for their precious comments and constructive suggestions, which contributed to the improvement of the quality of the paper. And then, the authors acknowledge the High-performance Computing Platform of Anhui University for providing computing resources.

\section*{Competing interests}
The authors declare that they have no competing interests.




\bibliographystyle{elsarticle-num-names} 
\bibliography{article}

\begin{thebibliography}{41}
\expandafter\ifx\csname natexlab\endcsname\relax\def\natexlab#1{#1}\fi
\providecommand{\url}[1]{\texttt{#1}}
\providecommand{\href}[2]{#2}
\providecommand{\path}[1]{#1}
\providecommand{\DOIprefix}{doi:}
\providecommand{\ArXivprefix}{arXiv:}
\providecommand{\URLprefix}{URL: }
\providecommand{\Pubmedprefix}{pmid:}
\providecommand{\doi}[1]{\href{http://dx.doi.org/#1}{\path{#1}}}
\providecommand{\Pubmed}[1]{\href{pmid:#1}{\path{#1}}}
\providecommand{\bibinfo}[2]{#2}
\ifx\xfnm\relax \def\xfnm[#1]{\unskip,\space#1}\fi
\bibitem[{Grysiewicz et~al.(2008)Grysiewicz, Thomas, and Pandey}]{1}
\bibinfo{author}{R.~A. Grysiewicz}, \bibinfo{author}{K.~Thomas},
  \bibinfo{author}{D.~K. Pandey},
\newblock \bibinfo{title}{Epidemiology of ischemic and hemorrhagic stroke:
  incidence, prevalence, mortality, and risk factors.},
\newblock \bibinfo{journal}{Neurologic Clinics} \bibinfo{volume}{26}
  (\bibinfo{year}{2008}) \bibinfo{pages}{871--895}.
\bibitem[{Yang et~al.(2019)Yang, Huang, Qi, Li, Liu, Wang, Zheng, and Wang}]{4}
\bibinfo{author}{H.~Yang}, \bibinfo{author}{W.~Huang}, \bibinfo{author}{K.~Qi},
  \bibinfo{author}{C.~Li}, \bibinfo{author}{X.~Liu}, \bibinfo{author}{M.~Wang},
  \bibinfo{author}{H.~Zheng}, \bibinfo{author}{S.~Wang},
\newblock \bibinfo{title}{Clci-net: Cross-level fusion and context inference
  networks for lesion segmentation of chronic stroke},
\newblock in: \bibinfo{booktitle}{International Conference on Medical Image
  Computing and Computer-Assisted Intervention},
  \bibinfo{organization}{Springer}, \bibinfo{year}{2019}, pp.
  \bibinfo{pages}{266--274}.
\bibitem[{Qi et~al.(2019)Qi, Yang, Li, Liu, Wang, Liu, and Wang}]{5}
\bibinfo{author}{K.~Qi}, \bibinfo{author}{H.~Yang}, \bibinfo{author}{C.~Li},
  \bibinfo{author}{Z.~Liu}, \bibinfo{author}{M.~Wang},
  \bibinfo{author}{Q.~Liu}, \bibinfo{author}{S.~Wang},
\newblock \bibinfo{title}{X-net: Brain stroke lesion segmentation based on
  depthwise separable convolution and long-range dependencies},
\newblock in: \bibinfo{booktitle}{International conference on medical image
  computing and computer-assisted intervention},
  \bibinfo{organization}{Springer}, \bibinfo{year}{2019}, pp.
  \bibinfo{pages}{247--255}.
\bibitem[{Xue et~al.(2020)Xue, Farhat, Boukrina, Barrett, Binder, Roshan, and
  Graves}]{6}
\bibinfo{author}{Y.~Xue}, \bibinfo{author}{F.~G. Farhat},
  \bibinfo{author}{O.~Boukrina}, \bibinfo{author}{A.~Barrett},
  \bibinfo{author}{J.~R. Binder}, \bibinfo{author}{U.~W. Roshan},
  \bibinfo{author}{W.~W. Graves},
\newblock \bibinfo{title}{A multi-path 2.5 dimensional convolutional neural
  network system for segmenting stroke lesions in brain mri images},
\newblock \bibinfo{journal}{NeuroImage: Clinical} \bibinfo{volume}{25}
  (\bibinfo{year}{2020}) \bibinfo{pages}{102118}.
\bibitem[{Qin et~al.(2018)Qin, Kamnitsas, Ancha, Nanavati, Cottrell, Criminisi,
  and Nori}]{7}
\bibinfo{author}{Y.~Qin}, \bibinfo{author}{K.~Kamnitsas},
  \bibinfo{author}{S.~Ancha}, \bibinfo{author}{J.~Nanavati},
  \bibinfo{author}{G.~Cottrell}, \bibinfo{author}{A.~Criminisi},
  \bibinfo{author}{A.~Nori},
\newblock \bibinfo{title}{Autofocus layer for semantic segmentation},
\newblock in: \bibinfo{booktitle}{International conference on medical image
  computing and computer-assisted intervention},
  \bibinfo{organization}{Springer}, \bibinfo{year}{2018}, pp.
  \bibinfo{pages}{603--611}.
\bibitem[{Liu et~al.(2020)Liu, Shahid, Sarapugdi, Lin, Chen, and Hua}]{8}
\bibinfo{author}{Y.-C. Liu}, \bibinfo{author}{M.~Shahid},
  \bibinfo{author}{W.~Sarapugdi}, \bibinfo{author}{Y.-X. Lin},
  \bibinfo{author}{J.-C. Chen}, \bibinfo{author}{K.-L. Hua},
\newblock \bibinfo{title}{Cascaded atrous dual attention u-net for tumor
  segmentation},
\newblock \bibinfo{journal}{Multimedia Tools and Applications}
  (\bibinfo{year}{2020}) \bibinfo{pages}{1--25}.
\bibitem[{Wang et~al.(2017)Wang, Jiang, Qian, Yang, Li, Zhang, Wang, and
  Tang}]{9}
\bibinfo{author}{F.~Wang}, \bibinfo{author}{M.~Jiang},
  \bibinfo{author}{C.~Qian}, \bibinfo{author}{S.~Yang},
  \bibinfo{author}{C.~Li}, \bibinfo{author}{H.~Zhang},
  \bibinfo{author}{X.~Wang}, \bibinfo{author}{X.~Tang},
\newblock \bibinfo{title}{Residual attention network for image classification},
\newblock in: \bibinfo{booktitle}{Proceedings of the IEEE conference on
  computer vision and pattern recognition}, \bibinfo{year}{2017}, pp.
  \bibinfo{pages}{3156--3164}.
\bibitem[{Dou et~al.(2020)Dou, Karimi, Rollins, Ortinau, Vasung, Velasco-Annis,
  Ouaalam, Yang, Ni, and Gholipour}]{10}
\bibinfo{author}{H.~Dou}, \bibinfo{author}{D.~Karimi}, \bibinfo{author}{C.~K.
  Rollins}, \bibinfo{author}{C.~M. Ortinau}, \bibinfo{author}{L.~Vasung},
  \bibinfo{author}{C.~Velasco-Annis}, \bibinfo{author}{A.~Ouaalam},
  \bibinfo{author}{X.~Yang}, \bibinfo{author}{D.~Ni},
  \bibinfo{author}{A.~Gholipour},
\newblock \bibinfo{title}{A deep attentive convolutional neural network for
  automatic cortical plate segmentation in fetal mri},
\newblock \bibinfo{journal}{IEEE transactions on medical imaging}
  \bibinfo{volume}{40} (\bibinfo{year}{2020}) \bibinfo{pages}{1123--1133}.
\bibitem[{Li et~al.(2021)Li, Li, and Fan}]{11}
\bibinfo{author}{Y.~Li}, \bibinfo{author}{H.~Li}, \bibinfo{author}{Y.~Fan},
\newblock \bibinfo{title}{Acenet: Anatomical context-encoding network for
  neuroanatomy segmentation},
\newblock \bibinfo{journal}{Medical Image Analysis} \bibinfo{volume}{70}
  (\bibinfo{year}{2021}) \bibinfo{pages}{101991}.
\bibitem[{Yu et~al.(2020)Yu, Wang, Gao, Yu, Shen, and Sang}]{26}
\bibinfo{author}{C.~Yu}, \bibinfo{author}{J.~Wang}, \bibinfo{author}{C.~Gao},
  \bibinfo{author}{G.~Yu}, \bibinfo{author}{C.~Shen},
  \bibinfo{author}{N.~Sang},
\newblock \bibinfo{title}{Context prior for scene segmentation},
\newblock in: \bibinfo{booktitle}{Proceedings of the IEEE/CVF Conference on
  Computer Vision and Pattern Recognition}, \bibinfo{year}{2020}, pp.
  \bibinfo{pages}{12416--12425}.
\bibitem[{Huang et~al.(2020)Huang, Shen, Chen, Wang, and Li}]{41}
\bibinfo{author}{J.~Huang}, \bibinfo{author}{H.~Shen},
  \bibinfo{author}{B.~Chen}, \bibinfo{author}{Y.~Wang},
  \bibinfo{author}{S.~Li},
\newblock \bibinfo{title}{Segmentation of paraspinal muscles at varied lumbar
  spinal levels by explicit saliency-aware learning},
\newblock in: \bibinfo{editor}{A.~L. Martel}, \bibinfo{editor}{P.~Abolmaesumi},
  \bibinfo{editor}{D.~Stoyanov}, \bibinfo{editor}{D.~Mateus},
  \bibinfo{editor}{M.~A. Zuluaga}, \bibinfo{editor}{S.~K. Zhou},
  \bibinfo{editor}{D.~Racoceanu}, \bibinfo{editor}{L.~Joskowicz} (Eds.),
  \bibinfo{booktitle}{Medical Image Computing and Computer Assisted
  Intervention -- MICCAI 2020}, \bibinfo{publisher}{Springer International
  Publishing}, \bibinfo{address}{Cham}, \bibinfo{year}{2020}, pp.
  \bibinfo{pages}{652--661}.
\bibitem[{Tang et~al.(2021)Tang, Yan, Cai, Huang, Xie, Xiao, Lu, Lin, and
  Lu}]{43}
\bibinfo{author}{Y.~Tang}, \bibinfo{author}{K.~Yan}, \bibinfo{author}{J.~Cai},
  \bibinfo{author}{L.~Huang}, \bibinfo{author}{G.~Xie},
  \bibinfo{author}{J.~Xiao}, \bibinfo{author}{J.~Lu}, \bibinfo{author}{G.~Lin},
  \bibinfo{author}{L.~Lu},
\newblock \bibinfo{title}{Lesion segmentation and recist diameter prediction
  via click-driven attention and dual-path connection},
\newblock in: \bibinfo{editor}{M.~de~Bruijne}, \bibinfo{editor}{P.~C. Cattin},
  \bibinfo{editor}{S.~Cotin}, \bibinfo{editor}{N.~Padoy},
  \bibinfo{editor}{S.~Speidel}, \bibinfo{editor}{Y.~Zheng},
  \bibinfo{editor}{C.~Essert} (Eds.), \bibinfo{booktitle}{Medical Image
  Computing and Computer Assisted Intervention -- MICCAI 2021},
  \bibinfo{publisher}{Springer International Publishing},
  \bibinfo{address}{Cham}, \bibinfo{year}{2021}, pp. \bibinfo{pages}{341--351}.
\bibitem[{Yu et~al.(2018)Yu, Wang, Peng, Gao, Yu, and Sang}]{60}
\bibinfo{author}{C.~Yu}, \bibinfo{author}{J.~Wang}, \bibinfo{author}{C.~Peng},
  \bibinfo{author}{C.~Gao}, \bibinfo{author}{G.~Yu}, \bibinfo{author}{N.~Sang},
\newblock \bibinfo{title}{Learning a discriminative feature network for
  semantic segmentation},
\newblock in: \bibinfo{booktitle}{2018 IEEE/CVF Conference on Computer Vision
  and Pattern Recognition}, \bibinfo{year}{2018}, pp.
  \bibinfo{pages}{1857--1866}. \DOIprefix\doi{10.1109/CVPR.2018.00199}.
\bibitem[{Fang et~al.(2019)Fang, Li, Pan, Li, and Yu}]{24}
\bibinfo{author}{C.~Fang}, \bibinfo{author}{G.~Li}, \bibinfo{author}{C.~Pan},
  \bibinfo{author}{Y.~Li}, \bibinfo{author}{Y.~Yu},
\newblock \bibinfo{title}{Globally guided progressive fusion network for 3d
  pancreas segmentation},
\newblock in: \bibinfo{booktitle}{International Conference on Medical Image
  Computing and Computer-Assisted Intervention},
  \bibinfo{organization}{Springer}, \bibinfo{year}{2019}, pp.
  \bibinfo{pages}{210--218}.
\bibitem[{Wang et~al.(2021)Wang, Cao, Feng, Xie, Yang, and Chen}]{25}
\bibinfo{author}{H.~Wang}, \bibinfo{author}{J.~Cao}, \bibinfo{author}{J.~Feng},
  \bibinfo{author}{Y.~Xie}, \bibinfo{author}{D.~Yang},
  \bibinfo{author}{B.~Chen},
\newblock \bibinfo{title}{Mixed 2d and 3d convolutional network with
  multi-scale context for lesion segmentation in breast dce-mri},
\newblock \bibinfo{journal}{Biomedical Signal Processing and Control}
  \bibinfo{volume}{68} (\bibinfo{year}{2021}) \bibinfo{pages}{102607}.
\bibitem[{Zhou et~al.(2019)Zhou, Huang, Dong, Xia, and Wang}]{31}
\bibinfo{author}{Y.~Zhou}, \bibinfo{author}{W.~Huang},
  \bibinfo{author}{P.~Dong}, \bibinfo{author}{Y.~Xia},
  \bibinfo{author}{S.~Wang},
\newblock \bibinfo{title}{D-unet: a dimension-fusion u shape network for
  chronic stroke lesion segmentation},
\newblock \bibinfo{journal}{IEEE/ACM transactions on computational biology and
  bioinformatics}  (\bibinfo{year}{2019}).
\bibitem[{Jie et~al.(2017)Jie, Li, Gang, and Albanie}]{14}
\bibinfo{author}{H.~Jie}, \bibinfo{author}{S.~Li}, \bibinfo{author}{S.~Gang},
  \bibinfo{author}{S.~Albanie},
\newblock \bibinfo{title}{Squeeze-and-excitation networks},
\newblock \bibinfo{journal}{IEEE Transactions on Pattern Analysis and Machine
  Intelligence} \bibinfo{volume}{PP} (\bibinfo{year}{2017}).
\bibitem[{Zhao et~al.(2017)Zhao, Shi, Qi, Wang, and Jia}]{17}
\bibinfo{author}{H.~Zhao}, \bibinfo{author}{J.~Shi}, \bibinfo{author}{X.~Qi},
  \bibinfo{author}{X.~Wang}, \bibinfo{author}{J.~Jia},
\newblock \bibinfo{title}{Pyramid scene parsing network},
\newblock in: \bibinfo{booktitle}{Proceedings of the IEEE conference on
  computer vision and pattern recognition}, \bibinfo{year}{2017}, pp.
  \bibinfo{pages}{2881--2890}.
\bibitem[{Gu et~al.(2019)Gu, Cheng, Fu, Zhou, Hao, Zhao, Zhang, Gao, and
  Liu}]{15}
\bibinfo{author}{Z.~Gu}, \bibinfo{author}{J.~Cheng}, \bibinfo{author}{H.~Fu},
  \bibinfo{author}{K.~Zhou}, \bibinfo{author}{H.~Hao},
  \bibinfo{author}{Y.~Zhao}, \bibinfo{author}{T.~Zhang},
  \bibinfo{author}{S.~Gao}, \bibinfo{author}{J.~Liu},
\newblock \bibinfo{title}{Ce-net: Context encoder network for 2d medical image
  segmentation},
\newblock \bibinfo{journal}{IEEE transactions on medical imaging}
  \bibinfo{volume}{38} (\bibinfo{year}{2019}) \bibinfo{pages}{2281--2292}.
\bibitem[{Gessert et~al.(2020)Gessert, Sentker, Madesta, Schmitz, Kniep,
  Baltruschat, Werner, and Schlaefer}]{44}
\bibinfo{author}{N.~Gessert}, \bibinfo{author}{T.~Sentker},
  \bibinfo{author}{F.~Madesta}, \bibinfo{author}{R.~Schmitz},
  \bibinfo{author}{H.~Kniep}, \bibinfo{author}{I.~Baltruschat},
  \bibinfo{author}{R.~Werner}, \bibinfo{author}{A.~Schlaefer},
\newblock \bibinfo{title}{Skin lesion classification using cnns with
  patch-based attention and diagnosis-guided loss weighting},
\newblock \bibinfo{journal}{IEEE Transactions on Biomedical Engineering}
  \bibinfo{volume}{67} (\bibinfo{year}{2020}) \bibinfo{pages}{495--503}.
  \DOIprefix\doi{10.1109/TBME.2019.2915839}.
\bibitem[{Wang et~al.(2020)Wang, Liu, Li, Xu, and Zhang}]{18}
\bibinfo{author}{G.~Wang}, \bibinfo{author}{X.~Liu}, \bibinfo{author}{C.~Li},
  \bibinfo{author}{Z.~Xu}, \bibinfo{author}{S.~Zhang},
\newblock \bibinfo{title}{A noise-robust framework for automatic segmentation
  of covid-19 pneumonia lesions from ct images},
\newblock \bibinfo{journal}{IEEE Transactions on Medical Imaging}
  \bibinfo{volume}{PP} (\bibinfo{year}{2020}) \bibinfo{pages}{1--1}.
\bibitem[{Woo et~al.(2018)Woo, Park, Lee, and Kweon}]{30}
\bibinfo{author}{S.~Woo}, \bibinfo{author}{J.~Park}, \bibinfo{author}{J.-Y.
  Lee}, \bibinfo{author}{I.~S. Kweon},
\newblock \bibinfo{title}{Cbam: Convolutional block attention module},
\newblock in: \bibinfo{booktitle}{Proceedings of the European conference on
  computer vision (ECCV)}, \bibinfo{year}{2018}, pp. \bibinfo{pages}{3--19}.
\bibitem[{Tang et~al.(2020)Tang, Yan, Xiao, and Summers}]{52}
\bibinfo{author}{Y.~Tang}, \bibinfo{author}{K.~Yan}, \bibinfo{author}{J.~Xiao},
  \bibinfo{author}{R.~M. Summers},
\newblock \bibinfo{title}{One click lesion recist measurement and segmentation
  on ct scans},
\newblock in: \bibinfo{booktitle}{MICCAI}, \bibinfo{year}{2020}.
\bibitem[{Akal et~al.(2020)Akal, Peng, and Hermosillo}]{46}
\bibinfo{author}{O.~Akal}, \bibinfo{author}{Z.~Peng},
  \bibinfo{author}{G.~Hermosillo},
\newblock \bibinfo{title}{Combonet: Combined 2d \& 3d architecture for aorta
  segmentation},
\newblock \bibinfo{journal}{ArXiv} \bibinfo{volume}{abs/2006.05325}
  (\bibinfo{year}{2020}).
\bibitem[{Wu et~al.(2021)Wu, Wang, Zhong, Lei, Wen, and Qin}]{49}
\bibinfo{author}{H.~Wu}, \bibinfo{author}{W.~Wang}, \bibinfo{author}{J.~Zhong},
  \bibinfo{author}{B.~Lei}, \bibinfo{author}{Z.~Wen}, \bibinfo{author}{J.~Qin},
\newblock \bibinfo{title}{Scs-net: A scale and context sensitive network for
  retinal vessel segmentation},
\newblock \bibinfo{journal}{Medical Image Analysis} \bibinfo{volume}{70}
  (\bibinfo{year}{2021}) \bibinfo{pages}{102025}. \URLprefix
  \url{https://www.sciencedirect.com/science/article/pii/S1361841521000712}.
  \DOIprefix\doi{https://doi.org/10.1016/j.media.2021.102025}.
\bibitem[{Wang et~al.(2020)Wang, Wu, Zhu, Li, Zuo, and Hu}]{48}
\bibinfo{author}{Q.~Wang}, \bibinfo{author}{B.~Wu}, \bibinfo{author}{P.~Zhu},
  \bibinfo{author}{P.~Li}, \bibinfo{author}{W.~Zuo}, \bibinfo{author}{Q.~Hu},
\newblock \bibinfo{title}{Eca-net: Efficient channel attention for deep
  convolutional neural networks},
\newblock in: \bibinfo{booktitle}{2020 IEEE/CVF Conference on Computer Vision
  and Pattern Recognition (CVPR)}, \bibinfo{year}{2020}, pp.
  \bibinfo{pages}{11531--11539}. \DOIprefix\doi{10.1109/CVPR42600.2020.01155}.
\bibitem[{Song et~al.(2022)Song, Du, Zhang, and Li}]{57}
\bibinfo{author}{Y.~Song}, \bibinfo{author}{X.~Du}, \bibinfo{author}{Y.~Zhang},
  \bibinfo{author}{S.~Li},
\newblock \bibinfo{title}{Two-stage segmentation network with feature
  aggregation and multi-level attention mechanism for multi-modality heart
  images},
\newblock \bibinfo{journal}{Computerized Medical Imaging and Graphics}
  \bibinfo{volume}{97} (\bibinfo{year}{2022}) \bibinfo{pages}{102054}.
  \URLprefix
  \url{https://www.sciencedirect.com/science/article/pii/S0895611122000271}.
  \DOIprefix\doi{https://doi.org/10.1016/j.compmedimag.2022.102054}.
\bibitem[{Ronneberger et~al.(2015)Ronneberger, Fischer, and Brox}]{12}
\bibinfo{author}{O.~Ronneberger}, \bibinfo{author}{P.~Fischer},
  \bibinfo{author}{T.~Brox},
\newblock \bibinfo{title}{U-net: Convolutional networks for biomedical image
  segmentation},
\newblock in: \bibinfo{booktitle}{International Conference on Medical image
  computing and computer-assisted intervention},
  \bibinfo{organization}{Springer}, \bibinfo{year}{2015}, pp.
  \bibinfo{pages}{234--241}.
\bibitem[{Oktay et~al.(2018)Oktay, Schlemper, Folgoc, Lee, Heinrich, Misawa,
  Mori, McDonagh, Hammerla, Kainz et~al.}]{34}
\bibinfo{author}{O.~Oktay}, \bibinfo{author}{J.~Schlemper},
  \bibinfo{author}{L.~L. Folgoc}, \bibinfo{author}{M.~Lee},
  \bibinfo{author}{M.~Heinrich}, \bibinfo{author}{K.~Misawa},
  \bibinfo{author}{K.~Mori}, \bibinfo{author}{S.~McDonagh},
  \bibinfo{author}{N.~Y. Hammerla}, \bibinfo{author}{B.~Kainz}, et~al.,
\newblock \bibinfo{title}{Attention u-net: Learning where to look for the
  pancreas},
\newblock \bibinfo{journal}{arXiv preprint arXiv:1804.03999}
  (\bibinfo{year}{2018}).
\bibitem[{SHI et~al.(2015)SHI, Chen, Wang, Yeung, Wong, and WOO}]{51}
\bibinfo{author}{X.~SHI}, \bibinfo{author}{Z.~Chen}, \bibinfo{author}{H.~Wang},
  \bibinfo{author}{D.-Y. Yeung}, \bibinfo{author}{W.-k. Wong},
  \bibinfo{author}{W.-c. WOO},
\newblock \bibinfo{title}{Convolutional lstm network: A machine learning
  approach for precipitation nowcasting},
\newblock in: \bibinfo{editor}{C.~Cortes}, \bibinfo{editor}{N.~Lawrence},
  \bibinfo{editor}{D.~Lee}, \bibinfo{editor}{M.~Sugiyama},
  \bibinfo{editor}{R.~Garnett} (Eds.), \bibinfo{booktitle}{Advances in Neural
  Information Processing Systems}, volume~\bibinfo{volume}{28},
  \bibinfo{publisher}{Curran Associates, Inc.}, \bibinfo{year}{2015}.
  \URLprefix
  \url{https://proceedings.neurips.cc/paper/2015/file/07563a3fe3bbe7e3ba84431ad9d055af-Paper.pdf}.
\bibitem[{Dou et~al.(2016)Dou, Chen, Jin, Yu, Qin, and Heng}]{50}
\bibinfo{author}{Q.~Dou}, \bibinfo{author}{H.~Chen}, \bibinfo{author}{Y.~Jin},
  \bibinfo{author}{L.~Yu}, \bibinfo{author}{J.~Qin}, \bibinfo{author}{P.-A.
  Heng},
\newblock \bibinfo{title}{3d deeply supervised network for automatic liver
  segmentation from ct volumes},
\newblock in: \bibinfo{editor}{S.~Ourselin}, \bibinfo{editor}{L.~Joskowicz},
  \bibinfo{editor}{M.~R. Sabuncu}, \bibinfo{editor}{G.~Unal},
  \bibinfo{editor}{W.~Wells} (Eds.), \bibinfo{booktitle}{Medical Image
  Computing and Computer-Assisted Intervention -- MICCAI 2016},
  \bibinfo{publisher}{Springer International Publishing},
  \bibinfo{address}{Cham}, \bibinfo{year}{2016}, pp. \bibinfo{pages}{149--157}.
\bibitem[{Zhang et~al.(2021)Zhang, Li, Du, Qin, Wang, Chen, Liu, Gao, Ma, and
  Lei}]{19}
\bibinfo{author}{Y.~Zhang}, \bibinfo{author}{H.~Li}, \bibinfo{author}{J.~Du},
  \bibinfo{author}{J.~Qin}, \bibinfo{author}{T.~Wang},
  \bibinfo{author}{Y.~Chen}, \bibinfo{author}{B.~Liu},
  \bibinfo{author}{W.~Gao}, \bibinfo{author}{G.~Ma}, \bibinfo{author}{B.~Lei},
\newblock \bibinfo{title}{3d multi-attention guided multi-task learning network
  for automatic gastric tumor segmentation and lymph node classification},
\newblock \bibinfo{journal}{IEEE Transactions on Medical Imaging}
  \bibinfo{volume}{40} (\bibinfo{year}{2021}) \bibinfo{pages}{1618--1631}.
\bibitem[{Chen et~al.(2017)Chen, Papandreou, Kokkinos, Murphy, and Yuille}]{16}
\bibinfo{author}{L.-C. Chen}, \bibinfo{author}{G.~Papandreou},
  \bibinfo{author}{I.~Kokkinos}, \bibinfo{author}{K.~Murphy},
  \bibinfo{author}{A.~L. Yuille},
\newblock \bibinfo{title}{Deeplab: Semantic image segmentation with deep
  convolutional nets, atrous convolution, and fully connected crfs},
\newblock \bibinfo{journal}{IEEE transactions on pattern analysis and machine
  intelligence} \bibinfo{volume}{40} (\bibinfo{year}{2017})
  \bibinfo{pages}{834--848}.
\bibitem[{Liew et~al.(2018)Liew, Anglin, Banks, Sondag, Ito, Kim, Chan, Ito,
  Jung, Khoshab et~al.}]{21}
\bibinfo{author}{S.-L. Liew}, \bibinfo{author}{J.~M. Anglin},
  \bibinfo{author}{N.~W. Banks}, \bibinfo{author}{M.~Sondag},
  \bibinfo{author}{K.~L. Ito}, \bibinfo{author}{H.~Kim},
  \bibinfo{author}{J.~Chan}, \bibinfo{author}{J.~Ito},
  \bibinfo{author}{C.~Jung}, \bibinfo{author}{N.~Khoshab}, et~al.,
\newblock \bibinfo{title}{A large, open source dataset of stroke anatomical
  brain images and manual lesion segmentations},
\newblock \bibinfo{journal}{Scientific data} \bibinfo{volume}{5}
  (\bibinfo{year}{2018}) \bibinfo{pages}{1--11}.
\bibitem[{Maier et~al.(2017)Maier, Menze, von~der Gablentz, H{\"a}ni, Heinrich,
  Liebrand, Winzeck, Basit, Bentley, Chen et~al.}]{56}
\bibinfo{author}{O.~Maier}, \bibinfo{author}{B.~H. Menze},
  \bibinfo{author}{J.~von~der Gablentz}, \bibinfo{author}{L.~H{\"a}ni},
  \bibinfo{author}{M.~P. Heinrich}, \bibinfo{author}{M.~Liebrand},
  \bibinfo{author}{S.~Winzeck}, \bibinfo{author}{A.~Basit},
  \bibinfo{author}{P.~Bentley}, \bibinfo{author}{L.~Chen}, et~al.,
\newblock \bibinfo{title}{Isles 2015-a public evaluation benchmark for ischemic
  stroke lesion segmentation from multispectral mri},
\newblock \bibinfo{journal}{Medical image analysis} \bibinfo{volume}{35}
  (\bibinfo{year}{2017}) \bibinfo{pages}{250--269}.
\bibitem[{Lin et~al.(2017)Lin, Goyal, Girshick, He, and Dollár}]{53}
\bibinfo{author}{T.-Y. Lin}, \bibinfo{author}{P.~Goyal},
  \bibinfo{author}{R.~Girshick}, \bibinfo{author}{K.~He},
  \bibinfo{author}{P.~Dollár},
\newblock \bibinfo{title}{Focal loss for dense object detection},
\newblock in: \bibinfo{booktitle}{2017 IEEE International Conference on
  Computer Vision (ICCV)}, \bibinfo{year}{2017}, pp.
  \bibinfo{pages}{2999--3007}. \DOIprefix\doi{10.1109/ICCV.2017.324}.
\bibitem[{Milletari et~al.(2016)Milletari, Navab, and Ahmadi}]{54}
\bibinfo{author}{F.~Milletari}, \bibinfo{author}{N.~Navab},
  \bibinfo{author}{S.-A. Ahmadi},
\newblock \bibinfo{title}{V-net: Fully convolutional neural networks for
  volumetric medical image segmentation},
\newblock in: \bibinfo{booktitle}{2016 Fourth International Conference on 3D
  Vision (3DV)}, \bibinfo{year}{2016}, pp. \bibinfo{pages}{565--571}.
  \DOIprefix\doi{10.1109/3DV.2016.79}.
\bibitem[{Guo et~al.(2021)Guo, Lei, Chen, Du, Frangi, Qin, Zhao, Shi, Xia, and
  Wang}]{55}
\bibinfo{author}{L.~Guo}, \bibinfo{author}{B.~Lei}, \bibinfo{author}{W.~Chen},
  \bibinfo{author}{J.~Du}, \bibinfo{author}{A.~F. Frangi},
  \bibinfo{author}{J.~Qin}, \bibinfo{author}{C.~Zhao},
  \bibinfo{author}{P.~Shi}, \bibinfo{author}{B.~Xia},
  \bibinfo{author}{T.~Wang},
\newblock \bibinfo{title}{Dual attention enhancement feature fusion network for
  segmentation and quantitative analysis of paediatric echocardiography},
\newblock \bibinfo{journal}{Medical Image Analysis} \bibinfo{volume}{71}
  (\bibinfo{year}{2021}) \bibinfo{pages}{102042}. \URLprefix
  \url{https://www.sciencedirect.com/science/article/pii/S1361841521000888}.
  \DOIprefix\doi{https://doi.org/10.1016/j.media.2021.102042}.
\bibitem[{Badrinarayanan et~al.(2017)Badrinarayanan, Kendall, and Cipolla}]{58}
\bibinfo{author}{V.~Badrinarayanan}, \bibinfo{author}{A.~Kendall},
  \bibinfo{author}{R.~Cipolla},
\newblock \bibinfo{title}{Segnet: A deep convolutional encoder-decoder
  architecture for image segmentation},
\newblock \bibinfo{journal}{IEEE Transactions on Pattern Analysis and Machine
  Intelligence} \bibinfo{volume}{39} (\bibinfo{year}{2017})
  \bibinfo{pages}{2481--2495}. \DOIprefix\doi{10.1109/TPAMI.2016.2644615}.
\bibitem[{{\c{C}}i{\c{c}}ek et~al.(2016){\c{C}}i{\c{c}}ek, Abdulkadir,
  Lienkamp, Brox, and Ronneberger}]{59}
\bibinfo{author}{{\"O}.~{\c{C}}i{\c{c}}ek}, \bibinfo{author}{A.~Abdulkadir},
  \bibinfo{author}{S.~S. Lienkamp}, \bibinfo{author}{T.~Brox},
  \bibinfo{author}{O.~Ronneberger},
\newblock \bibinfo{title}{3d u-net: Learning dense volumetric segmentation from
  sparse annotation},
\newblock in: \bibinfo{editor}{S.~Ourselin}, \bibinfo{editor}{L.~Joskowicz},
  \bibinfo{editor}{M.~R. Sabuncu}, \bibinfo{editor}{G.~Unal},
  \bibinfo{editor}{W.~Wells} (Eds.), \bibinfo{booktitle}{Medical Image
  Computing and Computer-Assisted Intervention -- MICCAI 2016},
  \bibinfo{publisher}{Springer International Publishing},
  \bibinfo{address}{Cham}, \bibinfo{year}{2016}, pp. \bibinfo{pages}{424--432}.
\bibitem[{Zhou et~al.(2018)Zhou, Siddiquee, Tajbakhsh, and Liang}]{29}
\bibinfo{author}{Z.~Zhou}, \bibinfo{author}{M.~M.~R. Siddiquee},
  \bibinfo{author}{N.~Tajbakhsh}, \bibinfo{author}{J.~Liang},
\newblock \bibinfo{title}{Unet++: A nested u-net architecture for medical image
  segmentation},
\newblock in: \bibinfo{booktitle}{Deep learning in medical image analysis and
  multimodal learning for clinical decision support},
  \bibinfo{publisher}{Springer}, \bibinfo{year}{2018}, pp.
  \bibinfo{pages}{3--11}.

\end{thebibliography}





\end{document}